\documentclass{article}
\usepackage{jfrExamplee}
\usepackage{graphicx}
\usepackage{setspace}
\usepackage{natbib}
\usepackage{subfiles}
\graphicspath{images/}
\usepackage{amsmath,amssymb}
\usepackage{amsfonts}
\usepackage{textcomp,gensymb}
\usepackage{algorithm}
\usepackage[dvipsnames]{xcolor}
\usepackage{hyperref}

\usepackage{epsfig}
\usepackage{epstopdf}
\usepackage{subfig}
\usepackage{xcolor}
\usepackage{authblk}
\usepackage{multirow}
\usepackage[noend]{algpseudocode}

\title{A vision-based robotic system for precision pollination of apples}

\author{
\textbf{Uddhav Bhattarai}$^1$, \textbf{Ranjan Sapkota}$^1$, \textbf{Safal Kshetri}$^1$, \textbf{Changki Mo}$^2$, \textbf{Matthew D. Whiting}$^3$, \textbf{Qin Zhang}$^1$, \textbf{Manoj Karkee}$^{1,*}$
}

\small{
\affil{$^1$Department of Biological Systems Engineering, Center for Precision and Automated Agricultural Systems, Washington State University, Prosser, WA, 99350, USA}
\affil{$^2$School of Mechanical and Materials Engineering, Washington State University Tri-Cities, Richland, WA, 99354, USA}
\affil{$^3$Department of Horticulture, Washington State University, Prosser, WA, 99350, USA}
\affil{$^*$Contact: manoj.karkee@wsu.edu}
}

\begin{document}

\maketitle
\renewcommand\Affilfont{\itshape\small}

\begin{abstract}
Global food production depends upon successful pollination, a process that relies on natural and managed pollinators. However, natural pollinators are declining due to factors such as climate change, habitat loss, and pesticide use. This paper presents an integrated robotic system for precision pollination in apples. The system consisted of a machine vision system to identify target flower clusters and estimate their positions and orientations, and a manipulator motion planning and actuation system to guide the sprayer to apply charged pollen suspension to the target flower clusters. The system was tested in the lab, followed by field evaluation in Honeycrisp and Fuji orchards. In the Honeycrisp variety, the robotic pollination system achieved a fruit set of 34.8\% of sprayed flowers with 87.5\% of flower clusters having at least one fruit when a 2 gm/l pollen suspension was used. In comparison, the natural pollination technique achieved a fruit set of 43.1\% with 94.9\% of clusters with at least one fruit. In Fuji apples, the robotic system achieved lower pollination success, with 7.2\% of sprayed flowers setting fruit and 20.6\% of clusters having at least one fruit, compared to 33.1\% and 80.6\%, respectively, with natural pollination. Fruit quality analysis showed that robotically pollinated fruits were comparable to naturally pollinated fruits in terms of color, weight, diameter, firmness, soluble solids, and starch content. Additionally, the system cycle time was 6.5 seconds per cluster. The results showed a promise for robotic pollination in apple orchards. However, further research and development is needed to improve the system and assess its suitability across diverse orchard environments and apple cultivars. \\

\end{abstract}

\section{Introduction}
Pollination plays a pivotal role ensuring global food supply, as it is crucial for producing a significant portion of fruits, vegetables, and nut crops around the world \citep{gallai2009guidelines}. In tree fruit production, growers rely on the pollinizer-pollinator model that involves planting compatible pollen-producing trees (pollinizers)  and renting beehives (pollinators) to enhance pollination and fruit set. However, the outcome of the pollinizer-pollinator model could be variable because of the variable pollinating capability of the pollinizer plant and environmental factors \citep{whiting2021supplemental}. Moreover, over the past few decades, the population of natural pollinators has been in a significant decline, primarily due to factors such as habitat loss, pesticide usage, climate change, and diseases \citep{national2007status,pollinatorthreatsnps}. Honeybees, a crucial pollinator for major crops, have been particularly affected by colony collapse disorder, a condition causing a rapid decline in beehives. A study estimates that the annual loss of bee colonies in the United States ranged from 40.4\% in 2018-19 to 44.0\% in 2019-20 \citep{bruckner2023national}. Given these circumstances, it is critical to develop efficient, scalable, and sustainable approach to pollination to ensure global food security.

To improve fruit and vegetable production, manual hand pollination has been in practice in small-scale farms, which involves the deposition of pollen particles by human labor using tools such as handheld brushes, sprayers, blowers, and vibrators \citep{wurz2021hand}. However, these methods are labor-intensive, costly, and challenging to scale, particularly in large-scale commercial orchards \citep{amador2017sticky}. In recent years, researchers have also investigated the design and fabrication of insect or bee-inspired micro-aerial vehicles that could be used for crop pollination \citep{wood2008first,yang2019bee+,dantu2011programming, ma2012design,colmenares2015compliant,chen2022design}. Some examples of such micro aerial vehicles include Robobee \citep{wood2008first}, Bee+\citep{yang2019bee+}, and Robofly \citep{chukewad2021robofly}. While these technological advancements are promising, coordinating swarms of robots at scale, addressing flight dynamics, control systems, and power supply remains a huge challenge. Furthermore, the precise maneuverability, agility, and adaptability of these technologies in outdoor environmental conditions for effective pollen delivery and crop pollination require further investigation and advancement.

Recent studies have also focused on using ground-based robotic systems and unmanned aerial vehicles (UAV) for pollinating fruits and vegetables such as tomatoes, brambles, kiwis,  vanilla, and dates \citep{shi2019study,yuan2016autonomous,ohi2018design, williams2020autonomous, li2022design}. Tomatoes and brambles are self-pollinating crops that rely on wind, vibration, and other motion-inducing agents to release pollens. For instance, to improve tomato wind pollination \citet{shi2019study} proposed modification of UAVs by adding air deflectors to alter downwash airflow direction. However, this modification negatively impacted flight stability, robustness, and control. In another work, \citet{ohi2018design} reported the autonomous robotic system ``BrambleBee", with a robotic manipulator attached to a ground vehicle for pollinating bramble flowers in greenhouses via contact-based interaction \citep{strader2019flower} to induce pollen release. In addition to tomato and bramble pollination in greenhouses, some studies are focused on ground-based robotic systems for kiwi pollination in outdoor environments with promising results \citep{williams2020autonomous, li2022design}.  \citet{williams2020autonomous} mounted a pollen spray system on an autonomous mobile platform developed by \citet{jones2019design} to navigate inside kiwi orchard. The spray system included a upward-facing stereo camera, an air-assisted spray manifold, and a kiwi flower identification system. Similarly, \citet{li2022design} designed and developed a six-degree-of-freedom serial robotic manipulator with stereo camera attached to the platform base and a sprayer attached at the distal end of the manipulator for kiwi pollination.
% The flower interaction system was equipped with linear servos for moving a cotton-padded end-effector plate that can face different directions to contact artificial bramble flowers . 

% Although Kiwis are not self-pollination, the kiwifruit trees are dioecious and have functional male and female flowers planted such that pollination could be achieved via insects or wind. For instance, 

% Some research efforts also focused on system development for pollinating tall trees such as date fruits and vanilla. \citet{shapiro2008robotic} developed a robotic platform with a two-degree-of-freedom (2 DOF) pan-tilt head attached on a telescopic mast for pollinating date fruit.\citet{shaneyfelt2013vision} proposed a crane system with two cranes attached at opposite side of docking post with each crane having a manipulator with a camera and sonar system for vanilla pollination. Researchers have also explored robotic systems for pollinating tall trees like date fruits and vanilla (Shapiro et al., 2008; Shaneyfelt et al., 2013). These systems typically involve robotic manipulators with cameras and sonar systems.

Despite recent advancements in pollination techniques, major challenges remain in pollinating cross-pollinating self-incompatible crops such as apples.  While apple flowers possess both male and female parts, they cannot self-pollinate and require cross-pollination by transferring pollens from compatible pollinizer trees. Hence, the vibratory and wind-based pollination approaches are not suitable for pollinating apples. Furthermore, unlike kiwi orchards, which are typically cultivated under overhead shade-like structures with kiwi flowers hanging from the canopy and generally no obstructions more than 20cm below the canopy \citep{williams2020autonomous},  apple flowers are randomly positioned in clusters facing random directions, presenting a significant challenge to effectively deposit pollen particles. Moreover, apple flowers are often obstructed by leaves, trunks, branches, trellis wires, and support posts in modern trellis-trained orchards, adding further complexity to the motion planning and manipulation.

This study introduced a ground-based robotic system for efficient and reliable pollination in cross-pollinating self-incompatible tree fruit crops (e.g., apples).  The robotic system performs a precision spray of pollens to the stigmatic surfaces of flowers using advanced machine learning algorithms, a six DOF robotic manipulator, and an electrostatic sprayer system. The system employed computer vision technology with an RGBD camera and deep learning algorithms to identify and delineate flower clusters. An electrostatic sprayer mechanism was utilized to maximize the pollen deposition, which is proven to enhance efficiency, coverage, and deposition compared to conventional non-electrostatic sprayer systems. Furthermore, the position and orientation of target flower clusters were computed in three-dimensional space, enabling strategic positioning of the sprayer nozzle for optimal pollen delivery. Field tests in commercial apple orchards, including two apple varieties (Honeycrisp and Fuji), demonstrated promising fruit-set results and fruit quality when compared to those achieved through conventional pollination techniques. The following are the main contributions of this work:
\begin{itemize}
    \item An integrated ground-based robotic system comprising a machine vision system, six-DOF robotic manipulation system, and an electrostatic sprayer system for precision pollination of tree fruit crops (e.g., apples). 
    \item Extensive field evaluation in two apple varieties (Honeycrisp and Fuji) that spanned multiple months encompassing various phases of fruit development, including pre-flowering netting, propagated pollen application during the flowering season, and subsequent assessment of fruit set and fruit quality using six distinct measures. 
    
\end{itemize}

\section{Robotic Pollination System Overview}
\begin{figure}[ht]
    \centering
    \includegraphics[width=0.99999\textwidth]{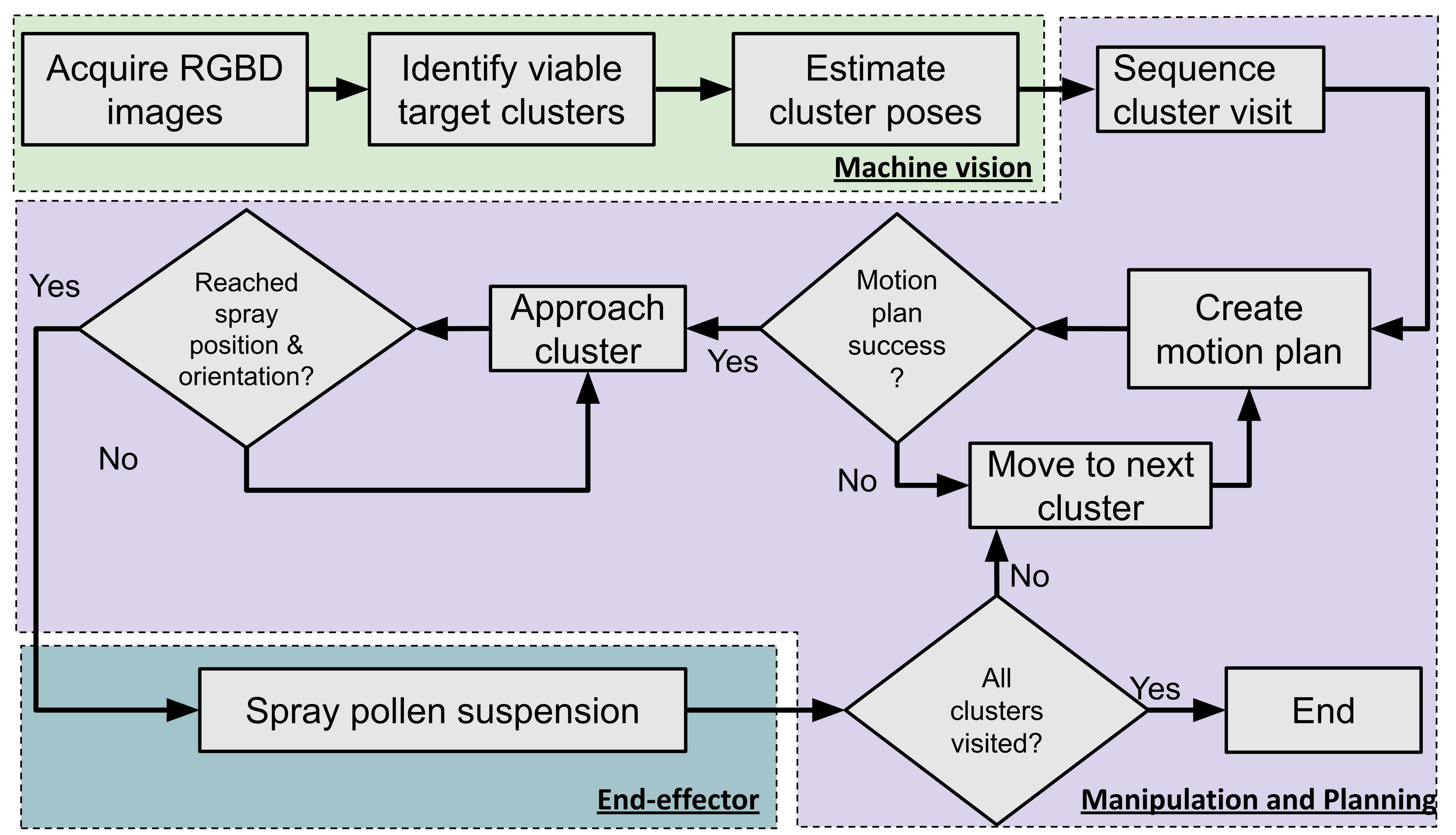}
    \caption{Overall flowchart of the proposed robotic pollination system. The system included a machine vision system with a RGB-D camera and image processing pipeline, and a mechatronic system with robotic manipulator, motion planning algorithms, and an electrostatic sprayer-based end-effector system.}
    \label{fig:overview}
\end{figure}
The proposed robotic pollination system comprised of three major components: i) a machine vision system to identify and locate the target flower clusters, ii) a robotic manipulator and motion planning system for collision-free navigation, iii) electrostatic sprayer-based end-effector system to spray positively charged pollen suspension to the target flower clusters (Figure \ref{fig:overview}). The system was initiated by acquiring an RGB image, which was processed by a Mask R-CNN-based deep learning algorithm to identify and segment target clusters in 2D image space. However, all segmented clusters may not be feasible targets due to manipulator workspace constraints, occlusion, or invalid depth information. Such invalid clusters were removed from the target list using an automated algorithm followed by a manual refinement by the operator. Once the viable target clusters were identified, the cluster position and orientation were estimated in 3D space to position the end-effector for effective pollen delivery. The viable target clusters were then automatically sequenced such that each cluster was visited only once while minimizing the manipulator travel costs. If a successful motion plan was computed, the electrostatic sprayer nozzle was navigated to a spray position 20cm away from the target cluster, facing the cluster surface orthogonally. Once the desired target position and orientation were reached, the pollen suspension was sprayed for 2 seconds, and the process was repeated until all the target clusters were visited.

\subsection{System Hardware}
\label{sec:system_hardware}

\begin{figure}[ht]
    \centering
    \includegraphics[width=0.48\textwidth]{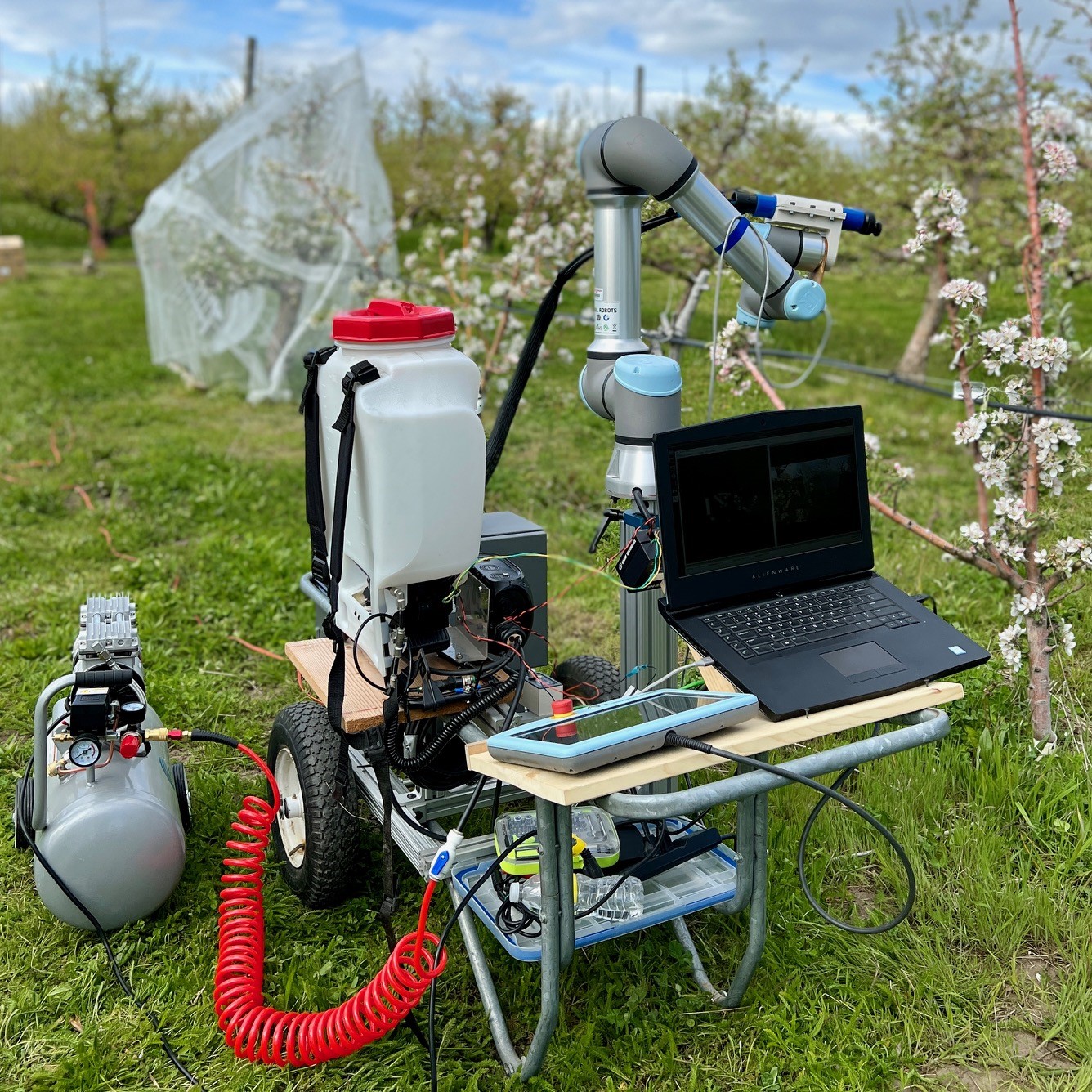}
    \includegraphics[width=0.48\textwidth]{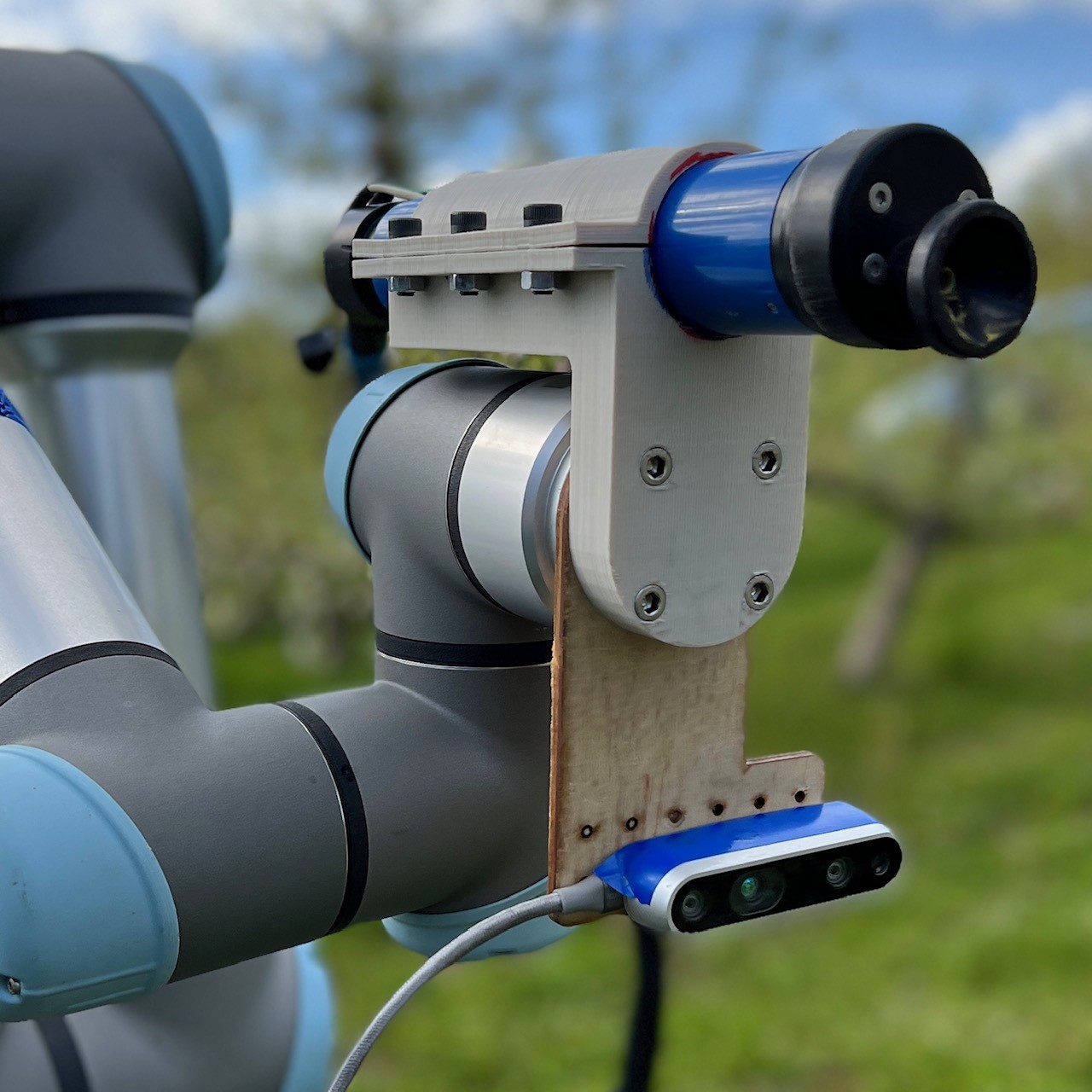}
    \caption{Robotic pollination system setup during the field trial in Naches and Pullman, WA. The processing laptop, machine vision, manipulation, and end-effector system were placed on a utility cart with an accompanying air compressor that supplied pressurized air for air-assisted atomization of pollen suspension. The electrostatic sprayer nozzle and the Intel RealSense D435i RGBD camera were rigidly attached to the distal end of the UR5e manipulator.}
    \label{fig:system_setup}
\end{figure}
\begin{figure}[ht]
    \centering
    \includegraphics[width=0.95\textwidth]{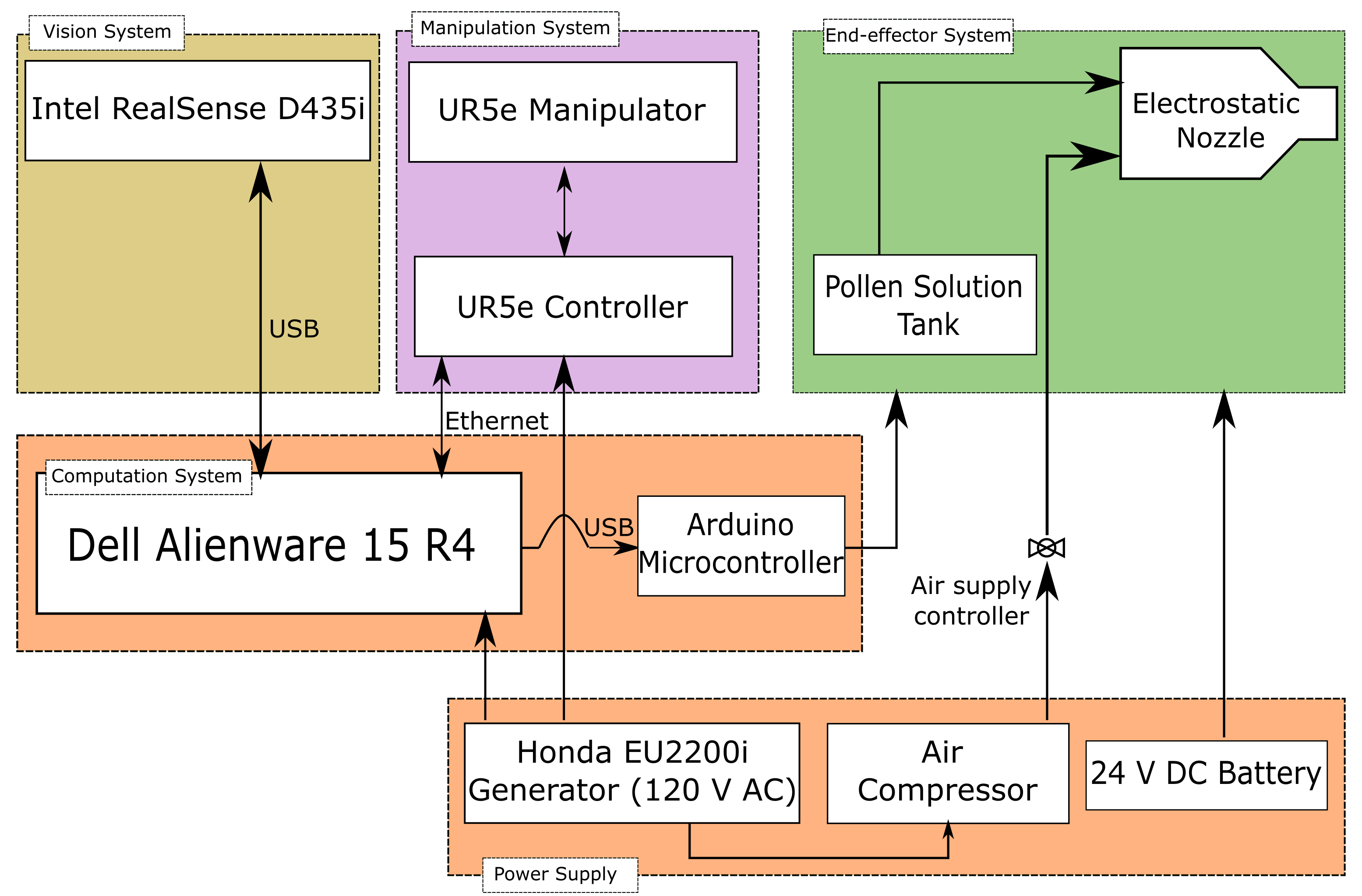}
    \caption{Hardware components of the proposed robotic pollination system. The components were categorized into machine vision system, manipulation system, end-effector system, computation system, and power supply.  Dell Alienware 15R4 laptop was utilized as a central control unit that handled processing, communication, and high-level control of the vision, manipulation, and end-effector systems.}
    \label{fig:hardware_setup}
\end{figure}

Figures \ref{fig:system_setup} and  \ref{fig:hardware_setup} illustrate the key hardware components of the proposed robotic pollination system that comprised machine vision system, manipulation system, electrostatic spayer-based end-effector system, computation system, and power supply. The machine vision system included an Intel RealSense D435i RGB-D camera as an eye-in-hand system with an imaging sensor to capture 2D RGB images and active stereo IR technology to estimate the depth. The RGB imaging sensor had an image resolution of $1920 \times 1080$ pixels and a Field of View (FOV) of  $69\degree  \times 42\degree$ whereas the depth estimation sensor had a resolution of $1280 \times 720$ pixels and a FOV of $87\degree \times 58\degree$. The mechatronic system included the UR5e robotic manipulator (Universal Robots, Odense, Denmark), manipulator controller, and an electrostatic sprayer system with a pollen suspension tank and electrostatic nozzle. For the sprayer system, the electrostatic sprayer mechanism was chosen instead of the conventional sprayer as it substantially improved spray efficiency and enhanced the coverage and deposition of sprayed particles \citep{whiting2021supplemental,das2016developing,law2001agricultural,badger2015electrostatic}. The required air supply for the spray pack was supplied from the external wheeled air compressor.  

All computational processing tasks were executed on a Dell Alienware 15R4 laptop, while the Arduino microcontroller was used for low-level control for the electrostatic sprayer (see Figure \ref{fig:hardware_setup}). The laptop had Ubuntu 18.04 operating system with ROS Melodic and included an Intel Core i7-8750 CPU, 2.2 GHz processor, 32 GB RAM, and an 8GB NVIDIA GeForce GTX 1080 Graphics Processing Unit (GPU). The manipulator was connected and controlled using a Local Area Network (LAN) between the laptop and the UR5e controller box. The Intel RealSense D435i camera and the Arduino microcontroller communicated via two USB connectors. The required power for the robotic manipulator, laptop, and air compressor unit was supplied by a Honda EU2200i (2200W, 120V AC, Honda Motor Co. Ltd, Tokyo, Japan) portable generator.  Furthermore, a 24 V battery attached at the base of the pollen suspension reservoir tank supplied the power for driving the electrostatic sprayer system, as shown in Figure \ref{fig:hardware_setup}. A wheeled mobile platform was used as the base for the installation of key hardware components, including the UR5e robotic manipulator, manipulator controller box, processing laptop, and spray pack, Figure \ref{fig:system_setup}. The eye-in-hand camera and the electrostatic sprayer nozzle were attached to the tool flange of the manipulator, allowing fixed transformation among the vision, manipulation, and end-effector coordinate frames. The robotic manipulator had a payload capacity of 5 KG and a reach of 850mm, enough to hold the camera and the sprayer nozzle.

\subsubsection{Electrostatic Sprayer System}
\begin{figure}[ht]
    \centering
    \includegraphics[width=0.9\textwidth]{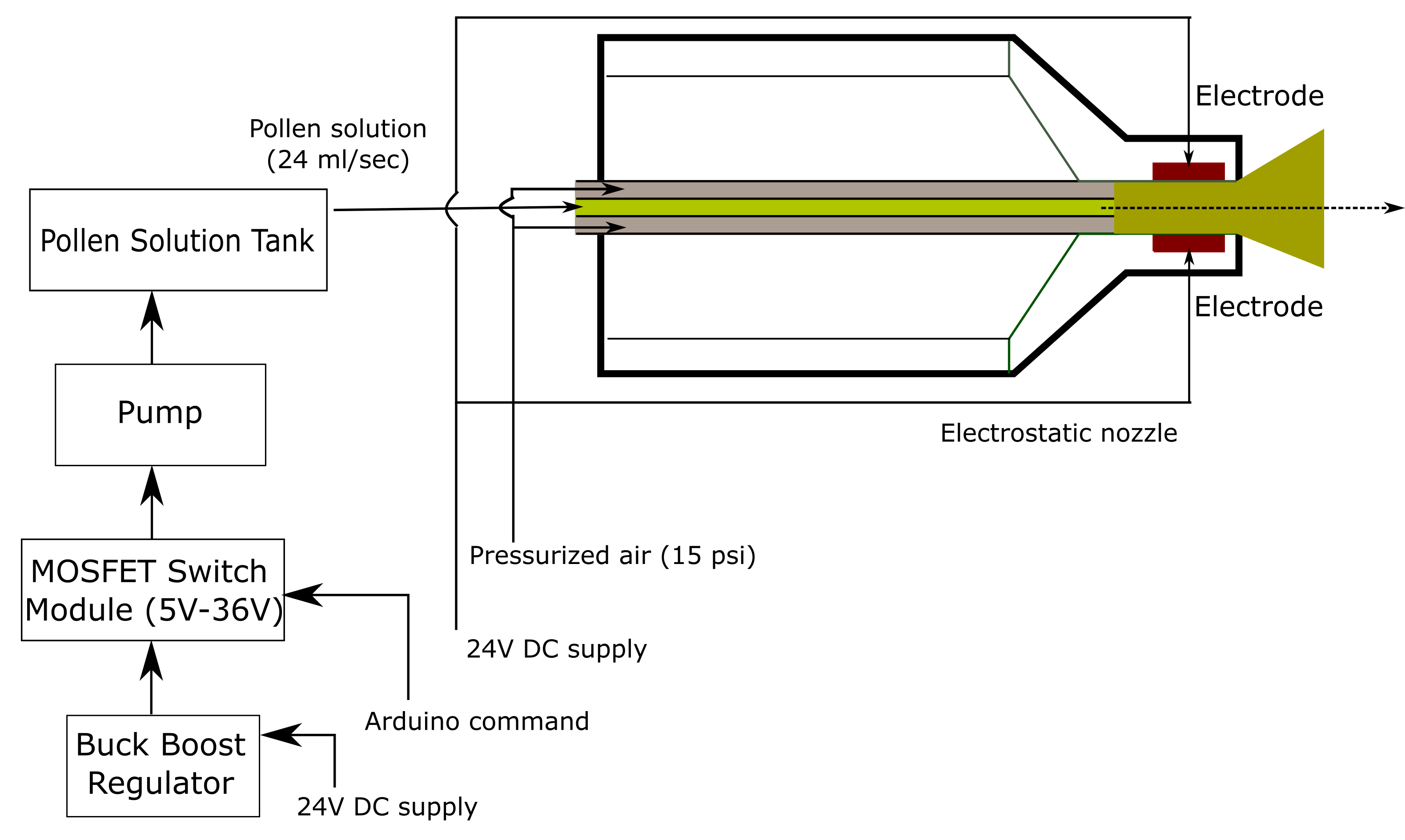}
  
    \caption{Electrostatic sprayer system used in this study. The pollen suspension was atomized through an air-assisted atomization process before being positively charged by the high-voltage electrode at the nozzle. The pump power supply was controlled to achieve the desired pollen flow rate.}
    \label{fig:electrostatic_sprayer}
\end{figure}
Figure \ref{fig:electrostatic_sprayer} illustrates the electrostatic sprayer system with an electrostatic charging mechanism and accompanying hardware components to operate the electrostatic nozzle. The liquid pollen suspension, pressurized air,  and supply voltage lines were connected to the sprayer nozzle. To create the pollen suspension, the cultivar-specific pollen powder was mixed and agitated with a proprietary suspension media developed and licensed by co-author Whiting \citep{whiting2021supplemental,das2016developing}, and water. The pollen suspension was atomized through air-assisted atomization, breaking down the pollen suspension into fine mists. When charged by electrostatic force, such mist particles repel each other, increasing the coverage area while being attracted by the grounded flower stigma,  significantly improving the chances of pollination and pollen germination. The pollen suspension supply was controlled by an electric pump attached to the bottom of the pollen suspension tank which also recirculated the pollen suspension, maintaining pollen suspension consistency during spray application. To regulate the pollen suspension flow rate, the pump power supply was varied by a Buck-Boost regulator that varied the input voltage. An Arduino microcontroller controlled the buck-boost regulator through a MOSFET switch, and the laptop controlled the microcontroller via serial connection.

% An electrostatic backpack sprayer system from OnTarget Spray Systems (Mt. Angel, Oregon) designed to conduct electrostatic spray was selected. The original spray pack was directly powered by a 24V DC power supply to control the pump to feed the pollen suspension and the electrostatic mechanism in the nozzle, providing a constant pollen suspension flow rate. This study utilized the modified version of the sprayer system to characterize the electrostatic spray output. The sprayer pack was modified by incorporating a buck-boost regulator to vary the supply voltage to the pump. The variable supply voltage to the pump allowed for fine-tuning the pollen suspension flow rate. Additionally, an Arduino microcontroller was added to actuate the sprayer pump and control the spray duration.  %In this study, the constant flow rate of 100mL/min was maintained by using a 24V supply voltage with a spray duration of 2 seconds.

An electrostatic backpack sprayer system from OnTarget Spray Systems (Mt. Angel, Oregon) was selected to conduct the electrostatic spray. A 24V DC power supply powered the original spray pack to control the pump to feed the pollen suspension and the electrostatic mechanism in the nozzle, providing a constant pollen suspension flow rate. This study utilized the modified version of the sprayer system, which incorporated a buck-boost regulator to vary the supply voltage \citep{ritchey2022characterization} to regulate the pollen suspension flow rate. Based on the preliminary experiments from \citet{ritchey2022characterization}, a constant flow rate of 1.573ml/sec was maintained using a 24V supply voltage with a spray duration of 2 seconds which was controlled by an Arduino microcontroller.

\subsection{Software Architecture}
\label{sec:software_architecture}
\begin{figure}[ht]
    \centering
    \includegraphics[width=0.95\textwidth]{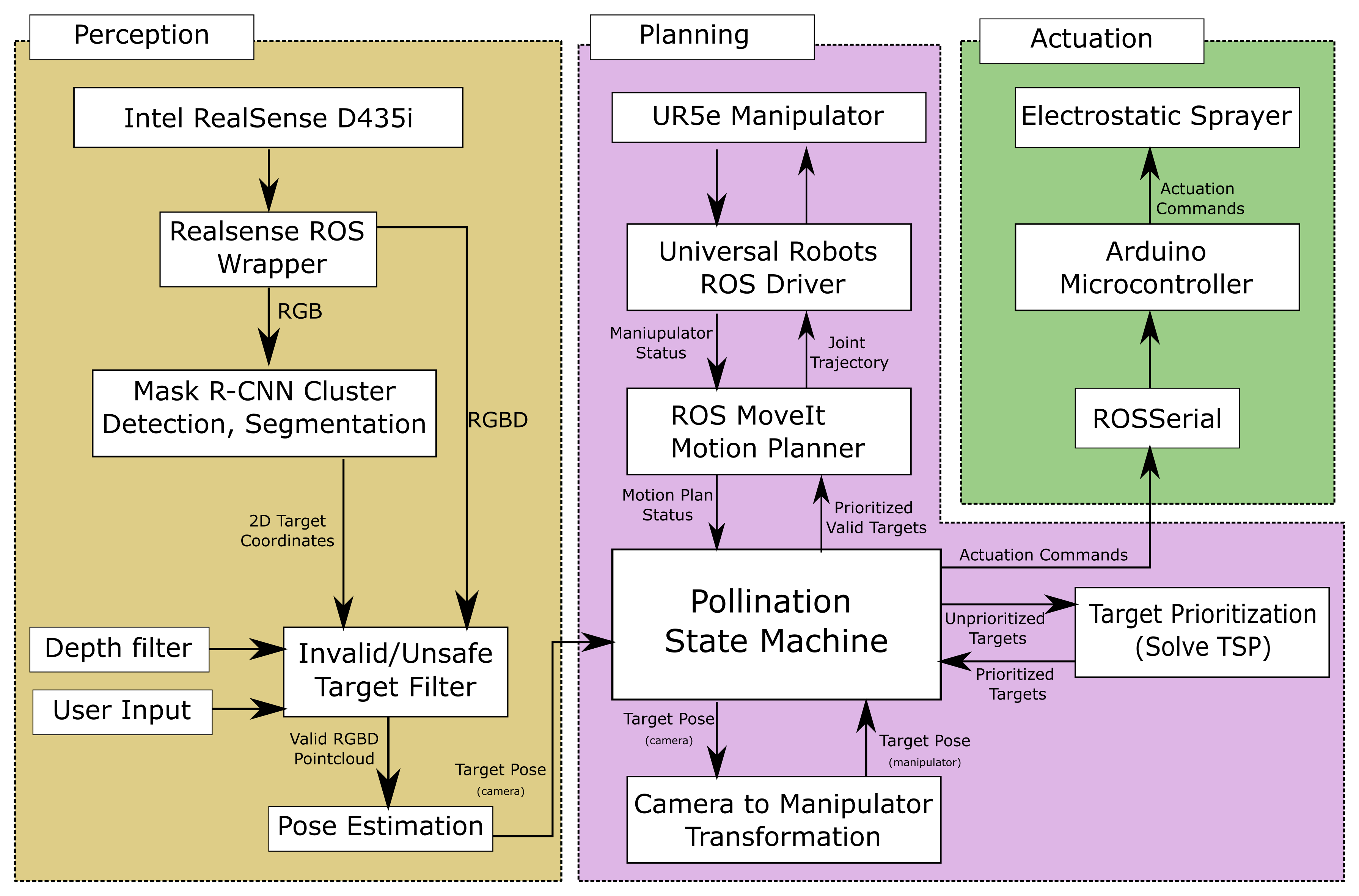}
    \caption{Software framework of the proposed robotic pollination system structured into three core components: machine vision, planning, and actuation. The machine vision system was responsible for tasks such as flower cluster identification, delineation, and pose estimation. A dedicated ``Pollination State Machine" was developed within the planning framework to gather relevant information from the machine vision system and facilitate the planning and navigation of the robotic manipulator and actuation of the end-effector.}
    \label{fig:software_architecture}
\end{figure}

The software architecture of the proposed robotic pollination system was divided into three distinct components: machine vision, planning, and actuation system, as illustrated in Figure \ref{fig:software_architecture}. The machine vision system comprised the pipeline for registered RGB-D data acquisition, flower cluster detection and segmentation, and localization in the camera coordinate space. The planning component was responsible for controlling overall system operation and coordination with the machine vision and the actuation system while managing the manipulator motion. The system coordination was achieved by creating an integrated state machine called the ``Pollination State Machine" within the planning framework. The actuation system included the electrostatic sprayer and relevant software modules to control the sprayer's actions.

\subsubsection{Machine Vision System}
The first step for conducting effective robotic pollination was to accurately identify and delineate apple flower clusters as possible pollination targets. To identify the target flower clusters, a Mask R-CNN model \citep{he2017mask} was modified, trained, and utilized \citep{bhattarai2020automatic}.  The instance segmentation capability of Mask R-CNN allowed precise cluster detection and boundary delineation for cluster pose (position and orientation) estimation in 3D space. The training image dataset consisted of 636 canopy images with 2,118 manually annotated flower clusters, randomly divided into training, validation, and test sets with a split of $70\%: 15\%: 15\%$ images. The Mask R-CNN algorithm used TensorFlow 2.0 with Keras as the backend engine and ResNet101 as the backbone. The network was trained up to 100 epochs using the Stochastic Gradient Descent (SGD) with a Learning Rate (LR) of 0.00005 and momentum of 0.9. The image dataset variability was further increased by combining different image augmentation techniques such as random horizontal and vertical flips, random scaling ($0.5$ -- $1.5$), and random rotation ($-60 \degree$ -- $+60\degree$).

The output of the instance segmentation algorithm was passed to the invalid/unsafe target filtering algorithm (Figure \ref{fig:software_architecture}). The target filtering algorithm employed two approaches for eliminating clusters that were considered invalid or unsafe. First, an automated approach eliminated clusters with invalid depth values, clusters facing the interior of the canopy, and clusters outside the manipulator working space with depth values of more than 1m. The second approach involved a manual safety check for removing unsafe clusters based on user input. Specifically, the clusters closer to the trellis wire, trunk, and support post that could potentially collide with the manipulator during the manipulator motion and spray operation were removed.  Once the valid target clusters were identified, their 3D positions and orientations were computed in camera space. To estimate the cluster's 3D position, 2D cluster RGB pixels were first projected onto 3D space utilizing depth information and camera intrinsic parameters, then averaging all the 3D cluster points. For estimating the cluster orientation, 3D cluster surface information was utilized. The apple flower cluster surface included both convex and concave surface structures. However, the majority of the cluster surfaces were found to be concave, with their surface normals pointing inward. Leveraging the concavity characteristic of the flower clusters, an algorithm was developed to estimate the clusters' orientation as discussed in our previous work \citep{bhattarai2024design}. Let $P_1 \dots P_N$ be the 3D position of all points in the flower cluster with camera viewpoint location $(v_p)$. To estimate the surface normal of the entire cluster, the cluster was divided into sub-sections defined by radius $(R)$ meter and the number of neighborhood points $(k)$. The surface normal of each cluster subsection was estimated by computing eigenvalues of the covariance matrix followed by normal refinement. The overall cluster orientation was estimated by averaging the refined surface normals \citep{bhattarai2024design}.

\begin{figure}[ht]
\centering
\includegraphics[scale=.25]{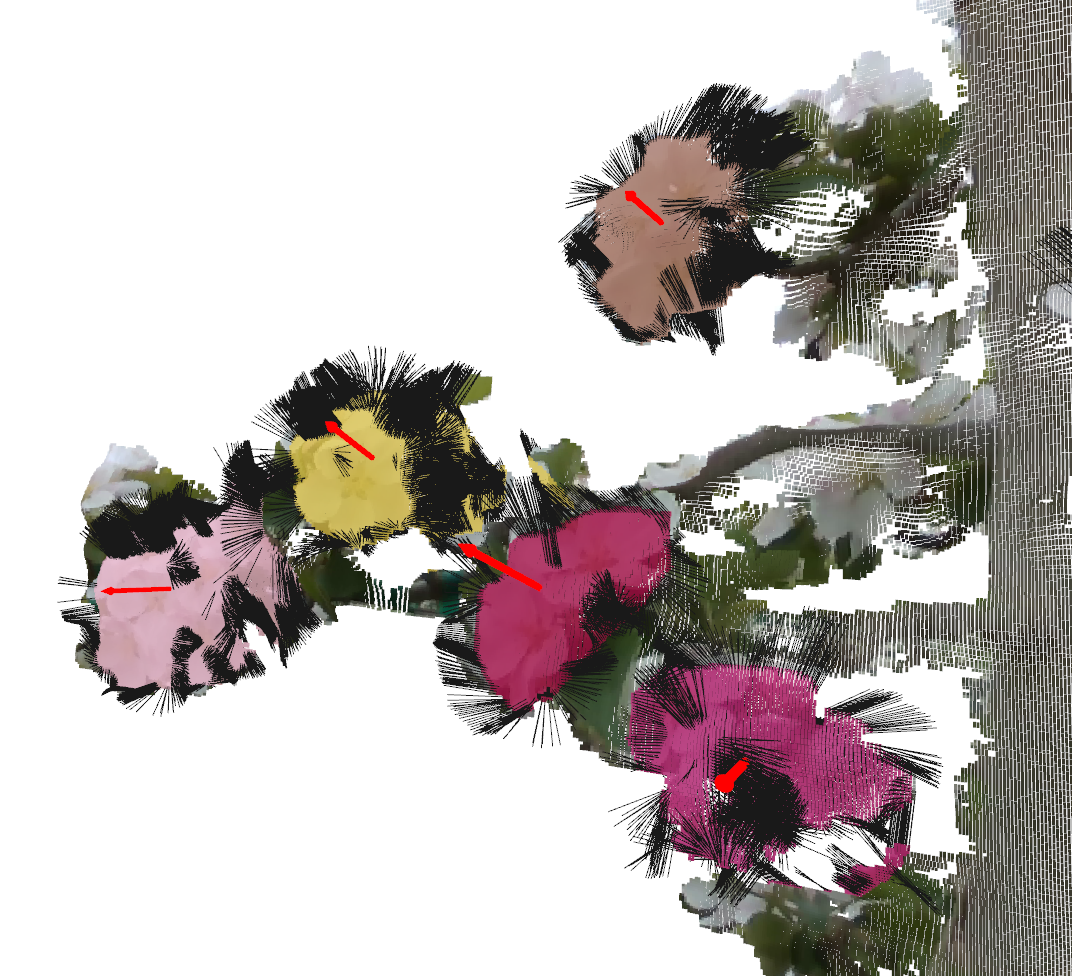}

\caption{
Representative 3D visualization of apple canopy after cluster segmentation and pose estimation. Different colored masks represent the apple flower clusters segmented by Mask R-CNN. Black lines represent the surface normals of each point of the segmented clusters. Most of the surface normal vectors face inward, representing the convexity of the cluster surface. Red arrows represent the estimated cluster position and orientation.}
\label{fig_chap6:clusterorientationestimation}
\end{figure}

Figure \ref{fig_chap6:clusterorientationestimation} shows an example of the position and orientation of the segmented flower clusters in the 3D canopy point cloud. Cluster instances obtained via the Mask R-CNN algorithm are reflected by different colored masks. The estimated cluster position and orientation are indicated by outward-facing red arrows, which also defined the approach angle of the end-effector. The surface normal estimated at every point $P_1 \dots P_N$ within the cluster is represented by black lines. Only the immediate front canopy is shown in the point cloud, as background canopies and objects were filtered out using depth filtering ($>$ 1m). Generally, the flowers were oriented such that estimated approach angles were generally outward-facing, which was likely due to two factors. First, the trellis-trained canopy caused the flowers to grow facing outside of the canopy. Second, the canopy was imaged using a front-facing camera with a single viewpoint for the pose estimation algorithm, which could have resulted in incomplete capture of the back sides of the clusters.

\subsubsection{Planning and Actuation}
Once the valid target cluster poses in the camera coordinate frame were obtained, they were transformed into the manipulator coordinate frame. Since the camera was rigidly attached to the tool flange of the robotic manipulator, there existed a fixed transformation between the manipulator and the camera coordinate frame. After the clusters were transformed into the manipulator coordinate space, the next step was to optimize the sequence of paths for the manipulator to visit individual clusters. For the given manipulator pose and target cluster poses, the objective was to identify the shortest feasible route for the manipulator such that the manipulator visited each cluster exactly once and returned to the start position, all while minimizing travel cost (Euclidean distance). This problem was modeled as the classic optimization problem, the Traveling Salesman Problem (TSP). To solve it, a Local Search algorithm was employed, utilizing OR-Tools, an open-source software designed for solving optimization problems \citep{googleortools}). 

For the manipulator motion planning, the ROS MoveIt package with an integrated Open Motion Planning Library (OMPL) was employed. The OMPL library was used to implement a motion plan, allowing the manipulator to find feasible paths and avoid obstacles in the environment. Based on the manipulator motion plan, joint trajectory commands were fed to the UR5e manipulator to navigate the electrostatic sprayer nozzle to the target position and orientation. Once the desired target pose was achieved, the actuation commands were fed to the Arduino microcontroller to actuate the sprayer. If the end-effector was successfully navigated to the target cluster and the electrostatically charged pollens suspension was successfully sprayed, target locations were updated, and the corresponding cluster was removed from the target list. Additionally, if the manipulator motion plan failed for a particular cluster and the spray was unsuccessful, the corresponding cluster was also removed from the target list. This process continued until all valid target clusters were processed.

\subsection{Laboratory Evaluation}
\begin{figure}[ht]
    \centering
    \includegraphics[width=0.305\textwidth]{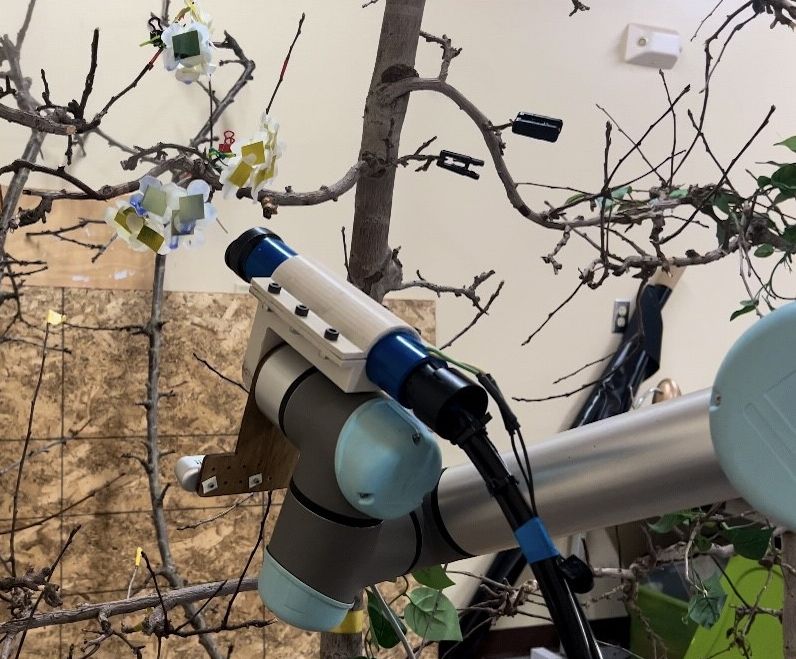}
    \includegraphics[width=0.335\textwidth]{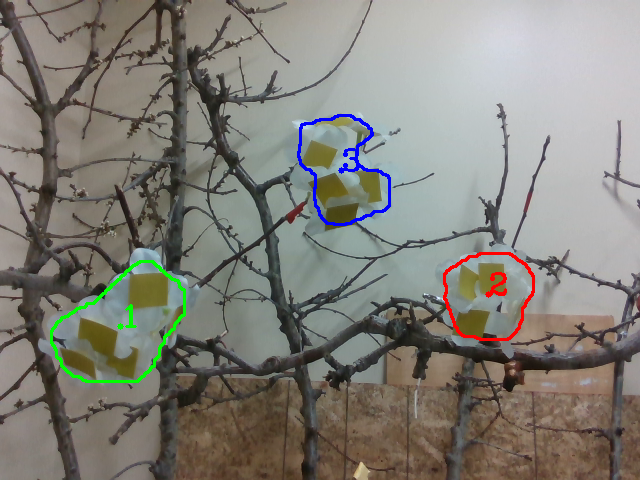}
    \includegraphics[width=0.32\textwidth]{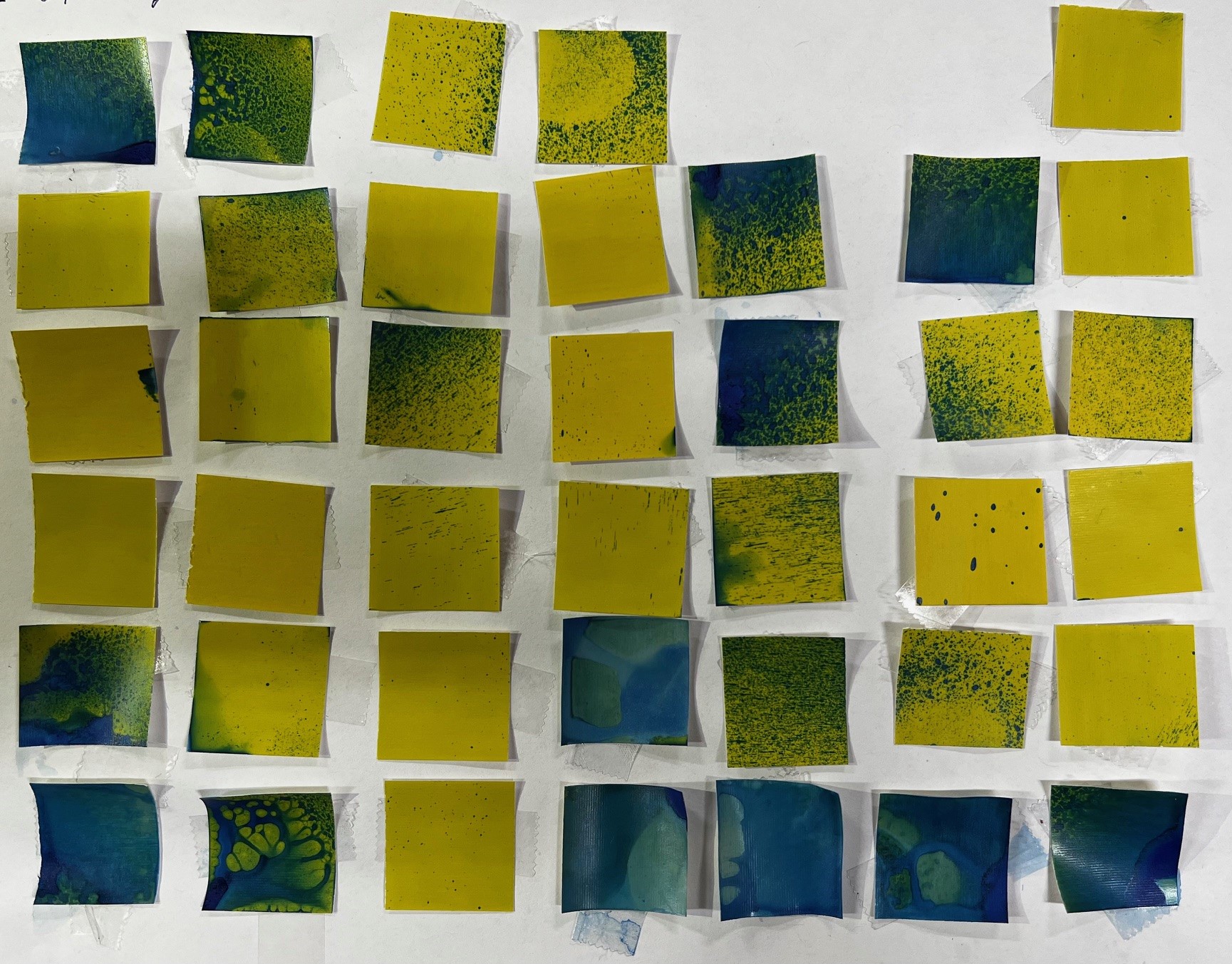}
    \caption{Laboratory evaluation of robotic pollination system. Manipulator in action during the experiment [Left], Mask-RCNN results in segmenting artificial flowers [Center], and collected water sensitive papers after the experiments [Right].}
    \label{fig:labexperiment}
\end{figure}

The integrated robotic system was evaluated in a laboratory environment for its ability to spray pollen suspension to target flowers. A laboratory-scale orchard was installed, and artificial flowers were attached at random positions and orientations within reach of the robotic arm (Figure \ref{fig:labexperiment}). Water-sensitive paper was cut into 1cm squares and attached at the center of each flower for spray coverage assessment.  The water-sensitive paper was coated with a chemical such as bromophenol blue, which changed color from yellow to blue in the presence of water. The system hardware and software were set up as described in sections \ref{sec:system_hardware} and \ref{sec:software_architecture}, and the experiments were conducted by spraying water to the target flower clusters. Figure \ref{fig:labexperiment} shows the robotic arm in action during the laboratory experiment [Left] and the segmentation results from the Mask R-CNN model [Center]. A total of 23 flower clusters consisting of an average of 5.65 water-sensitive papers per cluster were sprayed with water. Figure \ref{fig:labexperiment}[Right] shows the water-sensitive papers collected after the experiments, which shows the proportion of flower clusters being successfully sprayed by the robotic system. Each column represents flowers within a sprayed cluster. The water-sensitive paper changed to deep blue for flowers facing the spray direction.

\subsection{Experimental Evaluation in Apple Orchards}
% \subsection{Experimental Environment}
\begin{figure}[ht]
    \centering
    \includegraphics[width=0.245\textwidth]{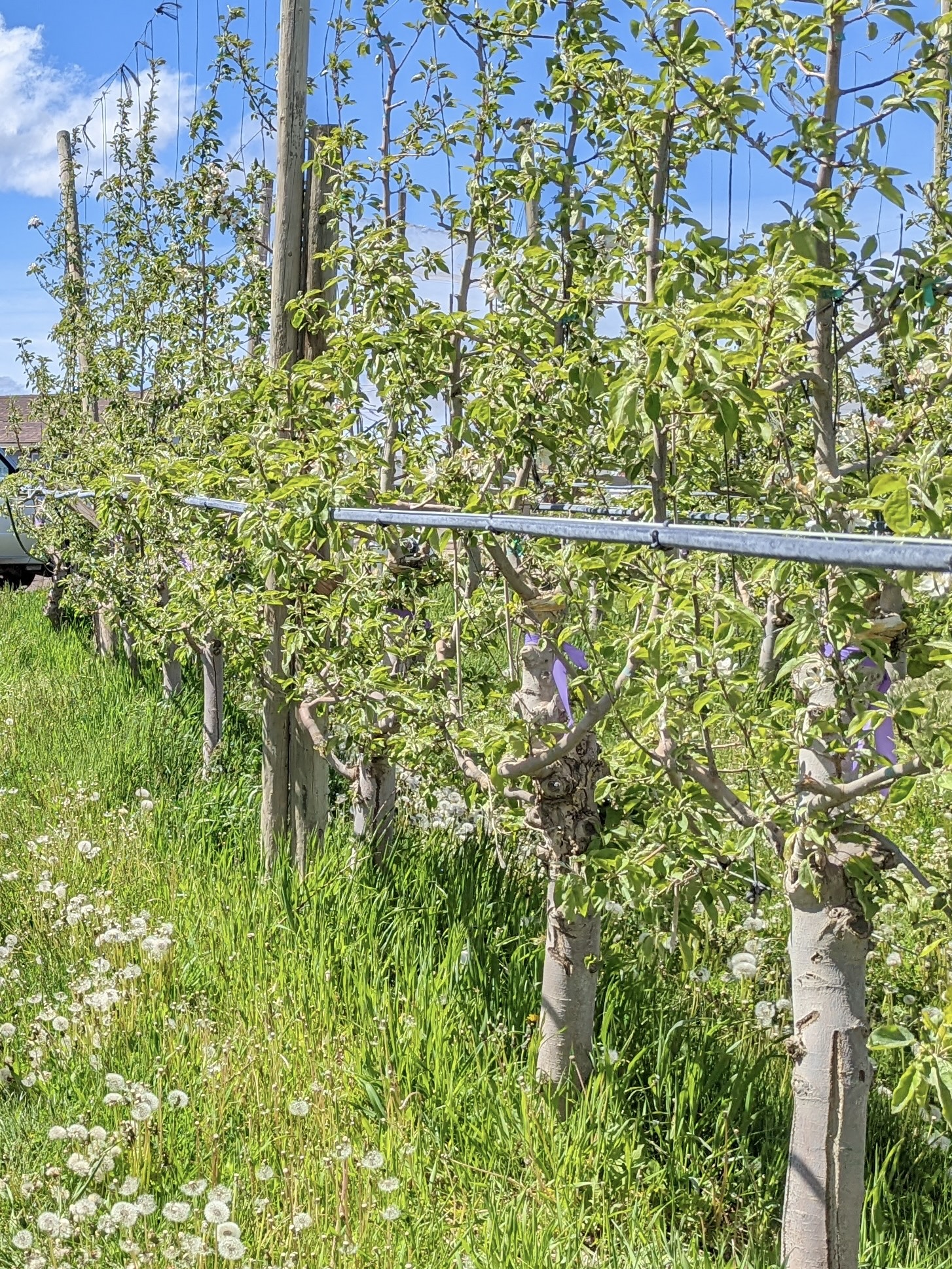}
    \includegraphics[width=0.245\textwidth]{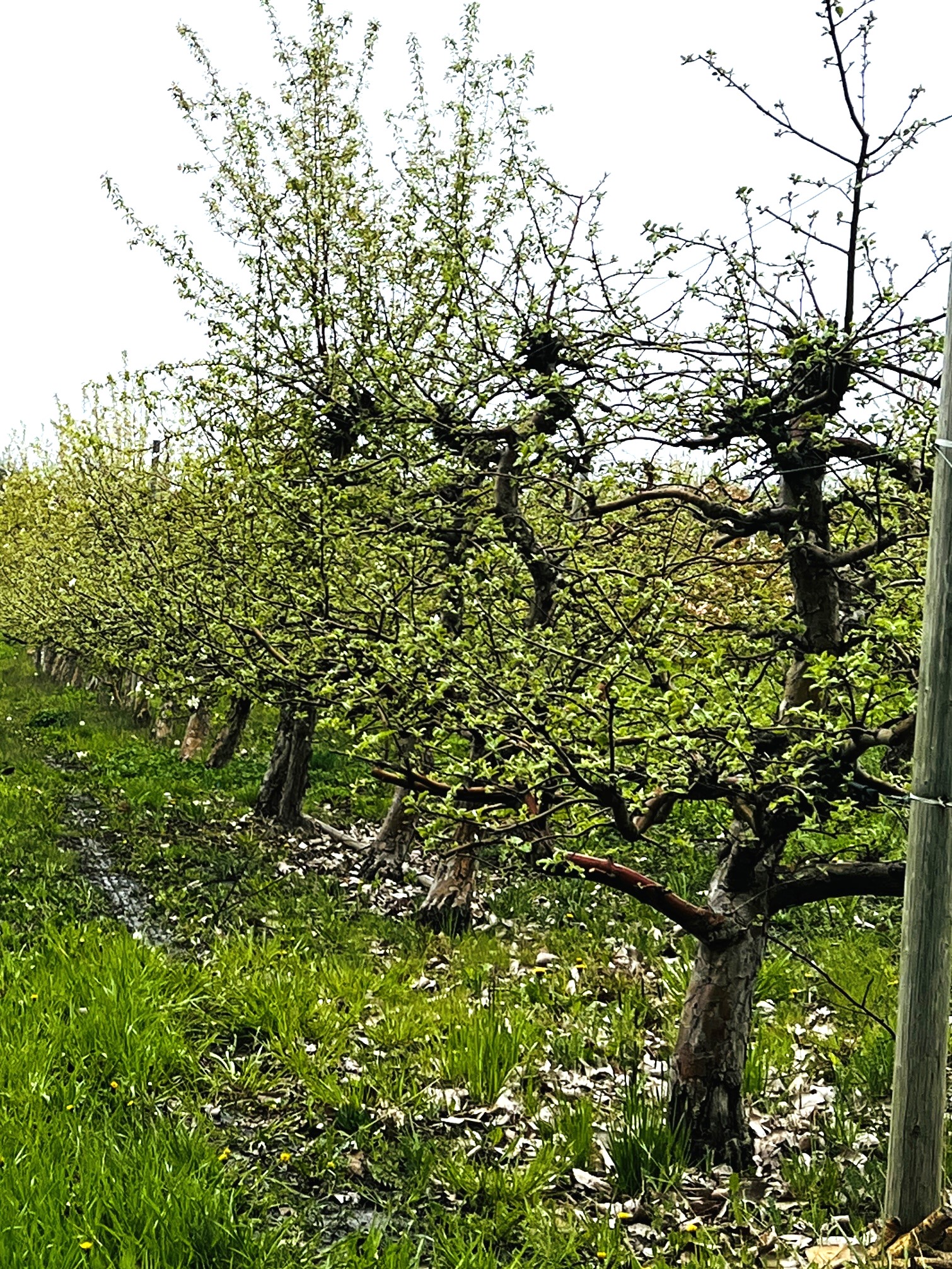}
    \includegraphics[width=0.48\textwidth]{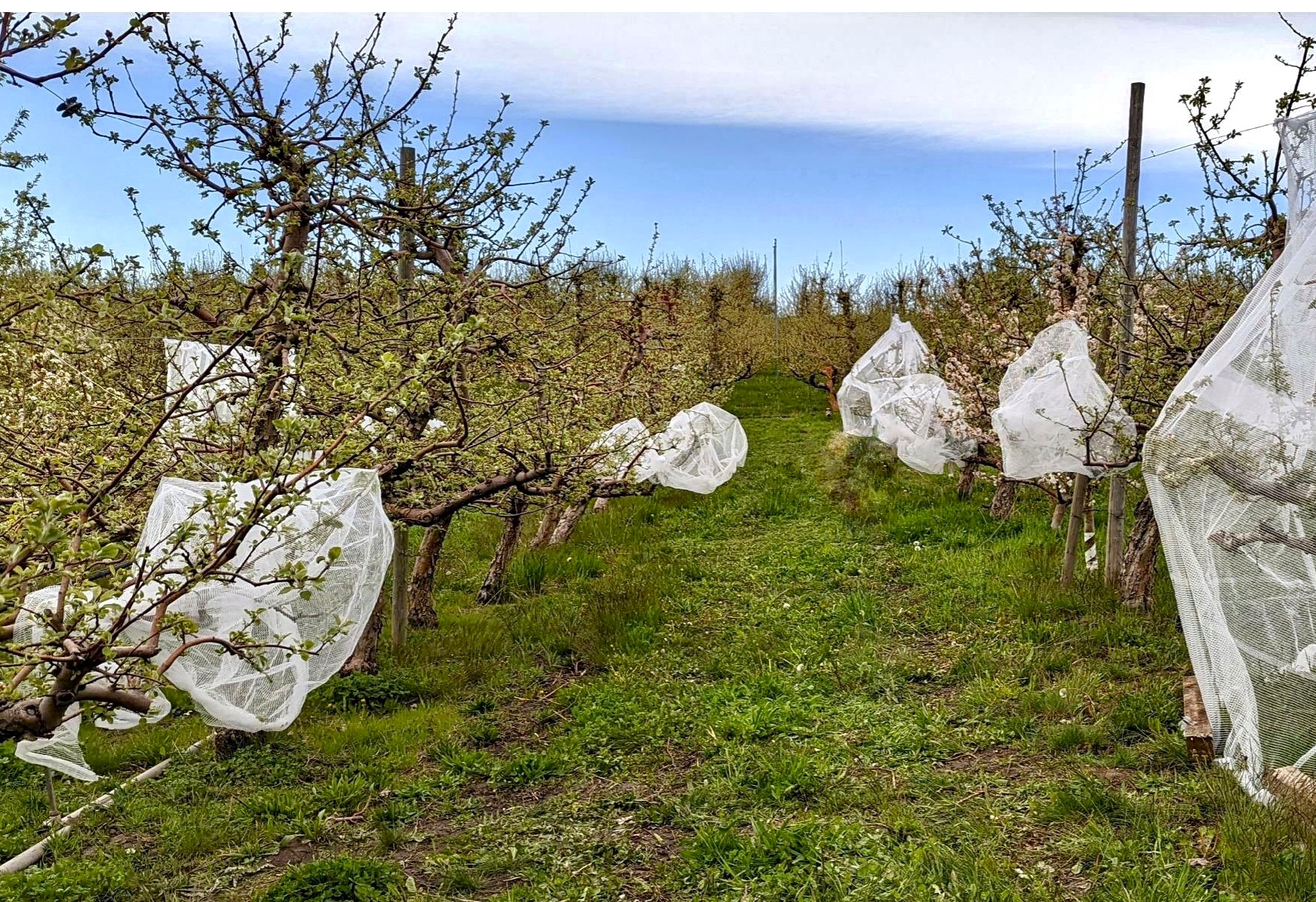}
    \caption{Experimental orchards for the field evaluation of prototype robotic pollination system: Honeycrisp orchard (Naches, WA) [Left]; Fuji orchard (Pullman, WA) [Center]. Examples of installed nets in canopy branches to avoid pollination via natural pollinators [Right].}
    \label{fig:experimental_orchards}
\end{figure}
To evaluate the performance of the prototype robotic pollination system in outdoor orchard environment, two experimental locations with different cultivars, geographical characteristics, and elevations were chosen (see Figure 8[Left, Center]. The first experimental site was a commercial ‘Honeycrisp’ apple orchard trained to a narrow vertical wall architecture in Naches, WA. The second was the Washington State University research orchard in Pullman, WA, where ‘Fuji’ trees were trained to a central leader system. The apple tree branches used in the experiment were netted before the budding flowers were open to prohibit pollination via natural pollinators such as bees and insects as shown in Figure \ref{fig:experimental_orchards} [Right]. The netting process involved creating an aluminum wire frame around the selected branches and then wrapping the net around the wireframe with securely fastened edges. During the field experiment, the nets were removed to spray the pollen and were immediately re-installed to avoid post-pollination pollinator activity. A mobile platform carrying the manipulator, controller box, spray pack, and processing laptop was manually pulled along the orchard rows along with the air compressor.

\begin{table}[h]
    \centering
    \begin{tabular}{lc}
        \hline \hline
         Parameters &  Specifications  \\
         \hline
         Pollen powder & 1 gm, 2 gm (Honeycrisp) \\
                        & 1 gm, 2 gm (Fuji) \\
         Pollen suspension media & 100 gm \\
         Water                   & 1 liter \\
         Pollen suspension replacement & Every 90 minutes \\
         Spray pressure             &15 psi \\
         Spray distance             &20 cm \\
         Spray time                 &2 seconds \\
         \hline \hline
    \end{tabular}
    \caption{Specifications of the different parameters used during the field trial of the proposed robotic pollination system.}
    \label{tab:specifications_details}
\end{table}

Table \ref{tab:specifications_details} summarizes experimental parameters used during the field trial. These experimental parameters were selected based on our preliminary laboratory experiments, as well as the experiments conducted by \cite{ritchey2022characterization}. Three different pollen suspension concentrations were developed by combining varying proportions (1 gm and 2 gm) of pollen particles with 100 gm pollen suspension media and one liter of water. The pollen suspension was carefully stirred and replaced every 90 minutes to improve pollen viability. Each target flower cluster was sprayed for 2 seconds from 20cm away while facing the cluster surface orthogonally. Throughout the experiment, the air compressor maintained a constant compressed air supply at 15 psi. The mobile platform remained stationary during the image acquisition and data processing and when the manipulator and end-effector were in motion. 

In Naches, a total of 82 flower clusters were evaluated and distributed across three pollination methods. There were 39 clusters for natural pollination, 16 clusters for robotic pollination with a pollen concentration of 2 gm/l, and 27 clusters for robotic pollination with a pollen concentration of 1 gm/l. Within these clusters, 378 individual flowers were assessed: 202 for natural pollination, 69 for robotic pollination with a 2 gm/l concentration, and 107 for robotic pollination with a 1 gm/l concentration.

At the Pullman site, the experiment involved 117 flower clusters of the Fuji variety. Among these, 62 clusters were subjected to natural pollination, 34 clusters to robotic pollination with a 2 gm/l pollen concentration, and 21 clusters to robotic pollination with a 1 gm/l concentration. These clusters contained a total of 542 individual flowers, with 284 undergoing natural pollination, 166 undergoing robotic pollination with a 2 gm/l concentration, and 92 undergoing robotic pollination with a 1 gm/l concentration. In both locations, the branches carrying the sprayed clusters were kept netted until five weeks after pollination. Once the blooming season was over and the fruit set was achieved, the netting was removed.

 % \begin{figure}
 %     \centering
 %     \includegraphics{Images/Orchards/NettedBranches_.jpg}
 %     \caption{Netted Branches}
 %     \label{fig:netted_branches}
 % \end{figure}

% \subsubsection{pollen suspension Development}
% To create the pollen suspension, the cultivar-specific pollen powder was mixed with a proprietary suspension media developed and privately licensed by co-author Whiting, and Water. Three different pollen suspensions were developed by combining varying proportions (0.5 gm, 1 gm, and 2 gm) of pollen particles with 100 gm pollen suspension media and one liter of water. The pollen suspension was carefully stirred and replaced every 90 minutes.

\section{System Evaluation}
\subsection{Machine Vision System Evaluation}
For each input image, the machine vision system generated a binary mask indicating whether each pixel belonged to the ``blossom" or ``background" class. The results of the machine vision algorithm were compared against the manually labeled blossom polygons, which served as the ground truth. Based on this comparison, the ``Precision" and ``Recall" metrics were computed, followed by the computation of the Average Precision (AP).
\begin{align}
    Average \: Precision (AP)=\sum_t (R_t-R_{t-1})P_t
\end{align}
where $R_t$ and $P_t$ are the precision and recall values for the classifier threshold of $t$. 

% Additionally, the manipulation system was evaluated by evaluating the capability of the manipulator to navigate the end-effector to the desired target location successfully out of all provided valid, sequenced candidate clusters selected for robotic pollination.
% \begin{align}
%     Manipulation \: Success \: (\%) =\frac{total \: sprayed \: clusters}{total \: valid \: target \: clusters} \times 100 \: (\%)
% \end{align}

\subsection{Cycle Time Evaluation}
The system cycle time was computed as the average time required for the robotic pollination system to complete one cycle of target flower cluster detection, segmentation, planning, navigation, and spraying. The machine vision system involved the time required to complete tasks such as image acquisition, Mask R-CNN instance segmentation, and 3D pose estimation, while the manipulation system included the time required for manipulator motion planning, target sequencing, and the time required for the physical motion of the manipulator to the desired target location.

\subsection{Fruit set and Fruit Quality Evaluation}
\label{FruitsetandQuality}
The performance of the proposed robotic pollination system was evaluated in two folds: fruit set and fruit quality evaluation. Fruits pollinated conventionally via natural pollinators such as bees and insects were used as the standard for evaluating the performance of the proposed robotic pollination system. Apple flowers that were in a similar state to the flowers that were sprayed for robotic pollination were randomly selected and tagged for the comparison. For the fruit set assessment, both apple orchard locations (Naches, WA; and Pullman, WA) were visited three weeks after the spray trial, and the fruit set was recorded. Two performance metrics were computed to evaluate the fruit set across different pollination treatments.
\begin{align}
    Fruit \: set \: Flowers (\%)=\frac{total \: fruitles}{total \: sprayed \: flowers} \times 100 (\%) \\
    Fruit \: set \: Clusters (\%)=\frac{total \: clusters \: with \: atleast \:one \:fruitlet}{total \: sprayed \: clusters} \times 100 (\%)
\end{align}

For fruit quality evaluation, fruits were harvested during the harvesting season, and the fruit quality was evaluated using external and internal quality metrics discussed below.

\begin{itemize}
    \item Fruit color (\% blush), weight (gm), and diameter (mm): The fruit color and size (weight and diameter) represent the fruit's appearance characteristics, often connected to the consumer's visual assessment of the fruit for purchase. The fruit color blush was evaluated on a scale of 0 to 100 \% by visually quantifying the surface area covered by the red blush. The fruit weight was measured using a digital measurement weight scale, while the diameter was measured along the equatorial line using a digital vernier caliper.

    \item  Firmness (lbf): The fruit firmness measures the force required to break the parenchyma cells in the fruit cortex \citep{musacchi2018apple}, and it relates to the consumer's perception of the fruit texture. Fruit firmness was measured using a penetrometer (GY-4 Digital Fruit Penetrometer) to determine the pressure required to puncture fruit flesh with peeled-off skin. Two different penetrometer readings were recorded by taking measurements on two opposite sides of the fruit cheek along the equatorial side, and the overall firmness was estimated by averaging the two measurements. 
    
    \item Soluble solid (\degree Brix):  During the fruit ripening, starch gets hydrolyzed into soluble carbohydrates and is used as a maturity indicator, with more starch being hydrolyzed, corresponding to increased fruit maturity \citep{mupambi2018influence,doerflinger2015variations}. Atago PAL-1 Digital Refractometer was used to measure the percentage of sucrose/soluble solid (\degree Brix) on the fruit. During the experiment, the fruit juice was squeezed on a refractometer, and the surface was rinsed with distilled water and wiped with soft fiber after each measurement.
    
    \item Starch: The starch-iodine test was also used to measure the fruit maturity. The fruit accumulates starch during the growth, which later converts into sugar once the fruit matures. For the starch-iodine test, iodine solution was evenly sprayed on the lower half of the fruit. Then, the sprayed fruits were left for 15 minutes, allowing them to create distinct patterns corresponding to varying levels of starch content. For Honeycrisp, the Starch Iodine index scale published by the Washington Tree Fruit Research Commission (WTFRC) \citep{WSTFRC} was used as a reference scale that categorized stained Honeycrisp apples on a scale of one to six, with a rating of six denoting ripe apples with minimal staining inside the fruit. For Fuji apples, the starch iodine index scale published by Cornell University \citep{blanpied1992predicting} was used, which measured apple maturity on a scale of one to eight.

    % \item Seed count: Seed count is a direct measure of the number of viable pollen grains that successfully fertilized the flower ovules. Previous research has suggested that the pollination method may affect the resulting seed count, a key indicator of pollination success \citep{samnegaard2019pollination, garratt2014avoiding}. To count the number of seeds, each apple was sliced transversely at the equator, and the number of seeds in the seed pockets was counted.

\end{itemize}
Parallel boxplots were used to visualize and compare the distribution of fruit quality metrics across different pollination treatments in two experimental orchards. Furthermore, the Kruskal-Wallis test, a nonparametric test for comparing medians of three or more groups, was used to assess the significance of differences in fruit quality among the pollination treatments, followed by posthoc Dunn's test for pairwise comparison of the pollination treatments.

\section{Results and Discussion}
\subsection{Machine Vision System}
\begin{figure}[ht]
\centering
\includegraphics[scale=.5]{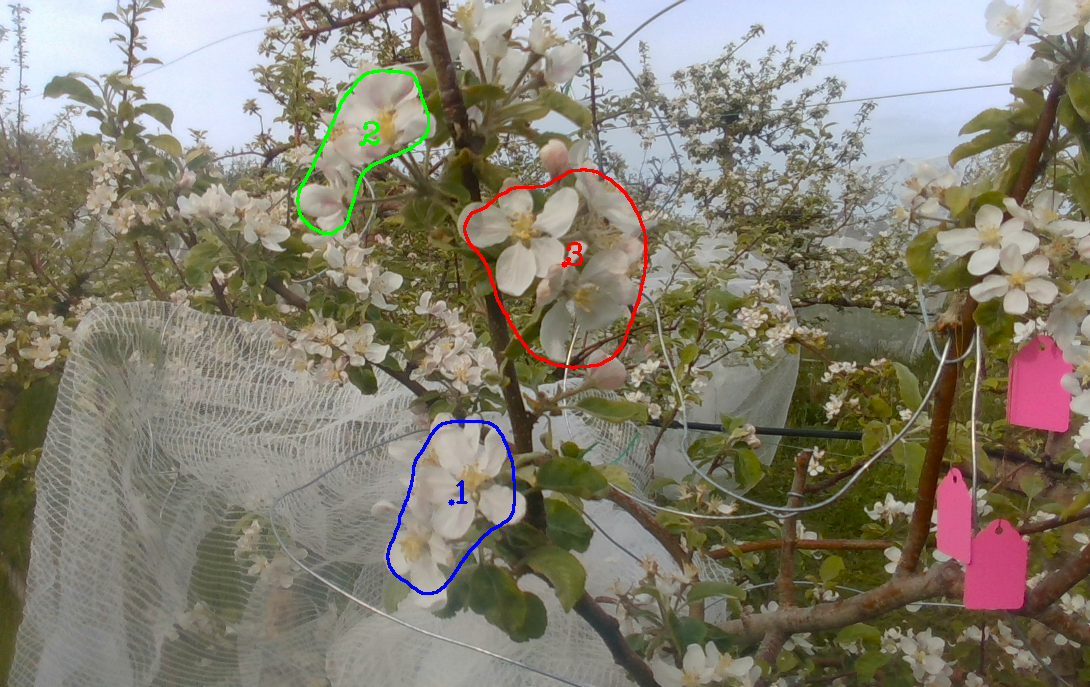}

\caption{Example flower cluster segmentation results achieved with a Mask RCNN model during the field evaluation in Pullman, WA. Tags were placed to identify each flower cluster after the application of a specific pollination treatment.}
\label{pullman_seg_result}
\end{figure}

The machine vision system achieved a mean average precision (mAP) of 0.89 in identifying and segmenting apple flower clusters in the test images. Figure \ref{pullman_seg_result} shows the segmentation results from the machine vision system during the field evaluation in Pullman, WA. As can be seen from the figure, the machine vision system correctly identified and segmented the apple flower clusters in the field images, even though the background varied significantly. The employed convolution neural network-based feature extraction was able to correctly learn feature information of the target flowers while discarding other canopy objects such as branches, leaves, trellis wire, flowers in the background, and the netting structure. Although the machine vision system showed promising results with a limited training dataset, the system's robustness and accuracy could be improved by adding more training images with variations in variety, canopy architectures, and lighting conditions. 

\subsection{Fruit Set with Robotic Pollination}

As shown in Table \ref{tab:fruitset_evaluation}, the robotic pollination approach with 2gm/l pollen concentration performed well, achieving a fruitset success of 34.8\% among the sprayed flowers, compared to the flowers treated with 1gm/l pollen concentration achieving a fruitset of 15.9\%. The better pollination success was also reflected in cluster-level pollination with 2gm/l pollen concentration, providing higher cluster-level fruitset success of 87.5\% compared to 37.0\% for 1gm/l pollen concentration. While the robotic pollination approaches performed slightly worse compared to the natural pollination (Fruit set Flowers: 43.1\%, Fruit set Clusters: 94.9\%), the achieved fruitset success is promising to offer a balanced cropload among the sprayed flowers. As seen in the results, about 87.5\% of the flower clusters successfully set at least one fruit, showing the good distribution of fruit set, which is desirable by the tree fruit growers. In tree fruit crops such as apples, growers perform crop load management approaches such as blossom and green fruit thinning for naturally pollinated fruits to reduce the crop load by 60 to 80\%. If all apple flowers are allowed to set fruits, it will result in a large number of small fruits with reduced fruit size and poor quality. Hence, with the fruitset of 34.8\% via robotic pollination, the effort of labor-intensive and costly blossom and green fruit thinning could be reduced and avoided. 
\begin{table}[ht]
    \centering
    \caption{Fruit set comparison of robotic vs natural Pollination for Honeycrisp and Fuji apple varieties at a flower and cluster level}
    \label{tab:fruitset_evaluation}
   \scalebox{0.85}{
   \begin{tabular}{ccccccc}
        \hline  \hline
        Experiment Site &Pollination Approach & Total Flowers &Fruit set Flowers (\%) &Total Clusters &Fruit set Clusters (\%) \\
        \hline
         Honeycrisp & Natural &202 &87 (43.1\%) &39 &37 (94.9\%) \\
         (Naches, WA) & Robot (2 gm/l) &69 &24 (34.8\%) &16 &14 (87.5\%) \\
          & Robot (1 gm/l) &107 &17 (15.9\%) &27 &10 (37.0\%) \\
        \hline
        Fuji & Natural &284 &94 (33.1\%) &62 &50 (80.6\%) \\
        (Pullman, WA) & Robot (2 gm/l) &166 &12 (7.2\%) &34 &7 (20.6\%) \\
         & Robot (1 gm/l) &92 &6 (6.5\%) &21 &3 (14.3\%) \\
         % & Robot (0.5 gm/l) &237 &19 (8.0\%) &53 &13 (24.5\%) \\
       
        \hline \hline
    \end{tabular}}
    
\end{table}

% with highest fruitset success of 43.1\% among the sprayed flowers with 94.9\% fruitset success in cluster level.
% In the Honeycrisp apple variety, the natural pollination approach achieved the highest fruit set (Fruit set Flowers: 43.1\%) followed by the robotic pollination approaches (see Table \ref{tab:fruitset_evaluation}). The robotic pollination approach with 2 gm/l (Fruit set Flowers: 36.8\%) pollen concentration was not as effective as the natural pollination, it performed well and showed better fruit set success compared to the flowers treated with 1 gm/l (Fruit set Flowers: 15.7\%) pollen concentration. The variation in pollination success was also reflected during the evaluation of cluster-level pollination. Natural pollination achieved the highest pollination success in the sprayed clusters (Fruit set Clusters: 94.9\%), followed by robotic pollination with pollen concentration of 2 gm/l (Fruit set Clusters: 36.8\%) and 1 gm/l (Fruit set Clusters: 15.7\%). 

Similar to the Honeycrisp apple variety,  the higher pollen concentration achieved a greater fruit set in Fuji apples with fruitset of 7.2\% for 2gm/l pollen concentration followed by 6.5\% for 1gm/l pollen concentration. The fruitset of both robotic pollination approaches was substantially lower than the natural pollination which had the fruitset success of 33.1\% .   At the cluster level as well, natural pollination performed the best (Fruit set Clusters: 80.6\%), followed by 2 gm/l (Fruit set Clusters: 20.6\%), and 1 gm/l (Fruit set Clusters: 14.3\%) pollen concentration.  It's worth noting that there was a noticeable decrease in pollination success for both flower- and cluster-level pollination in the Fuji variety compared to the Honeycrisp variety in all pollination approaches, with robotic pollination being particularly affected (see Table \ref{tab:fruitset_evaluation}).
\begin{figure}[ht]
    \centering
    \includegraphics[width=0.45\textwidth]{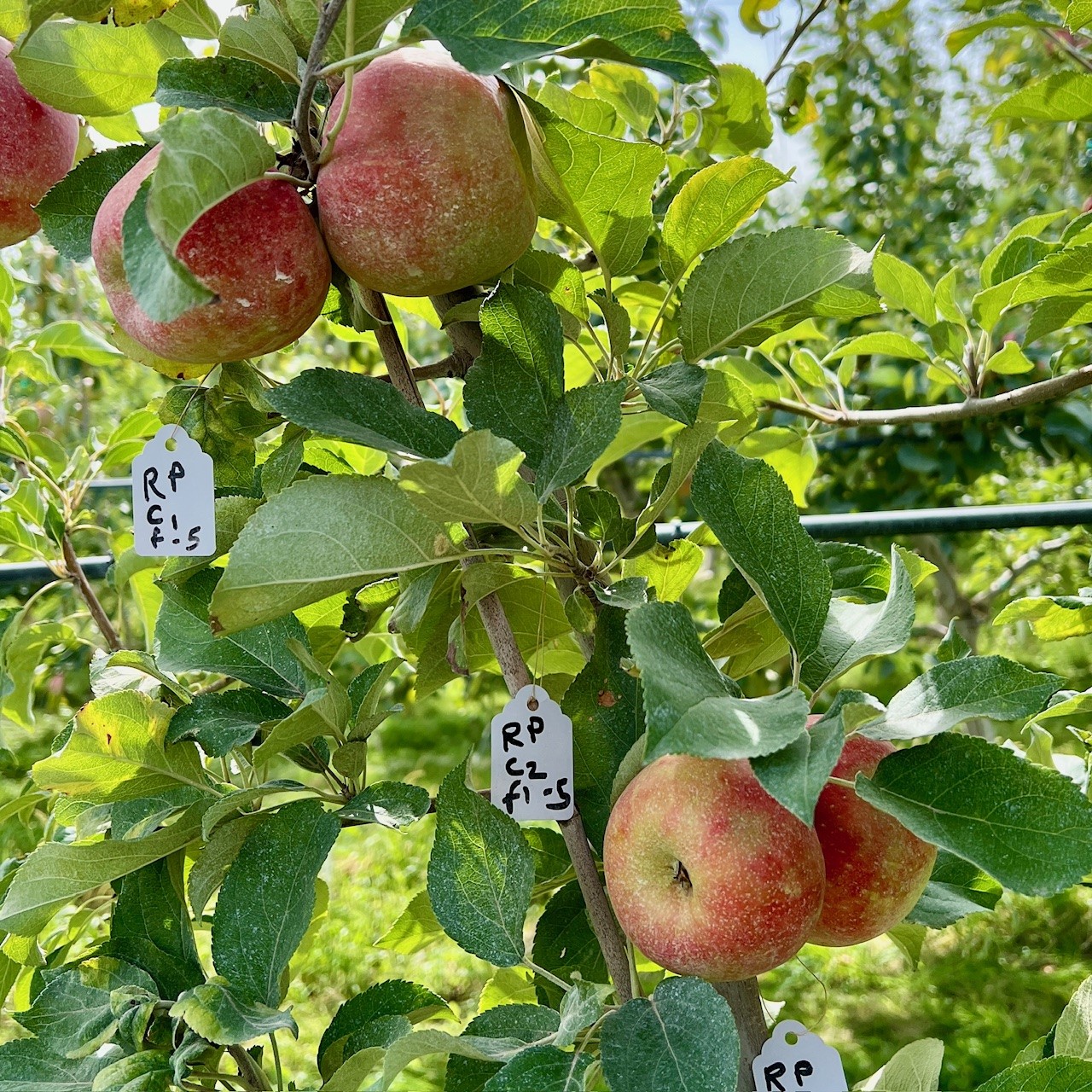}
    \includegraphics[width=0.515\textwidth]{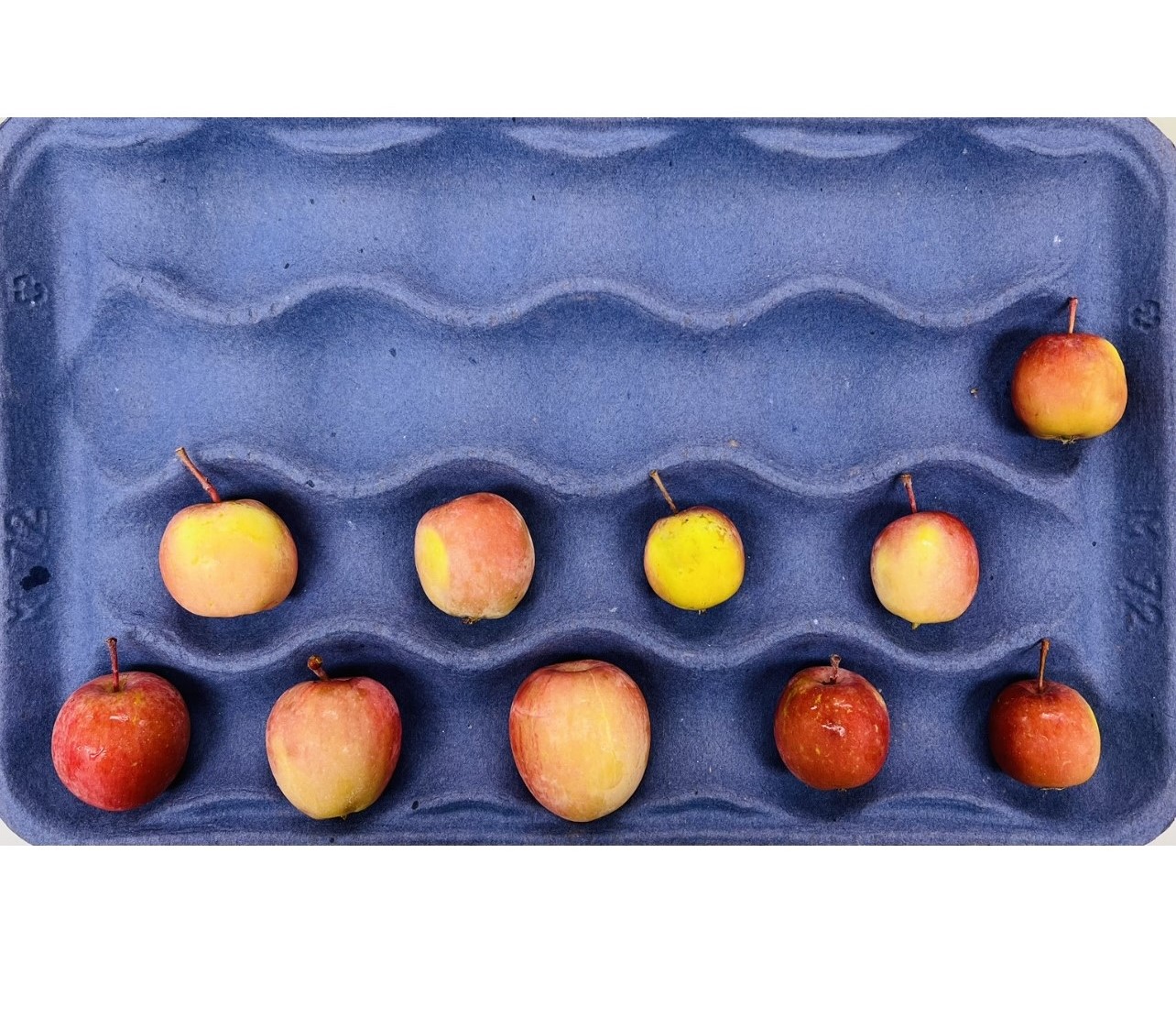}
    \caption{ Harvest-ready Honeycrisp apples pollinated via robotic pollination with 1 gm/l pollen concentration [Left], Harvested Fuji apples pollinated via robotic pollination with 2 gm/l pollen concentration during the fruit quality evaluation [Right]}
    \label{fig:pollination_harvested}
\end{figure}

The results showed that robotic pollination for both Honeycrisp and Fuji apple varieties was less effective in achieving fruitset compared to natural pollination. However, robotic pollination may still be a viable alternative, especially when pollen concentrations were high because only a small fraction of the flowers needs to be successfully pollinated to achieve 
 the desired fruit set that leads to optimal fruit yield and quality. The lower pollination success of robotic pollination may be caused by different factors, including the efficiency of pollen delivery, spray duration, spray timing relative to blooming, the quality of the pollen, and the environmental conditions during and after pollen spray.  Additionally, natural pollinators were highly likely to visit flowers multiple times over the entire bloom period, substantially increasing the chances of successful pollination. On the other hand, the robotic pollination approach in this study utilized a single spray cycle with a single opportunity to germinate pollen particles.

 The inter-variety comparison of fruit set showed that Fuji apples had a lower fruit set compared to Honeycrisp apples in both natural and robotic pollination approaches. The results of this study are in contrast to those of a previous study, which found that Honeycrisp apples having significantly lower fruit set (66.67\% Honeycrisp vs. 80\% for Fuji) and pollen germination rate than Fuji apples \citep{jahed2017pollen} for natural pollination. However, it is important to note that pollen germination success can be affected by the age of the canopy, pollen source, temperature, and other environmental factors \citep{zebro2023effects}. The Honeycrisp orchard in Naches, WA, was newly established compared to the Fuji apple orchard in Pullman, WA, which was established in the late 1980s. Some research suggests that the pollination capabilities of apple canopies may diminish with increasing age. Furthermore, especially during our field trials, the weather conditions may have a more pronounced negative impact on fruit set in Fuji apples. While both locations had mostly clear skies with intermittent clouds during the trial, Fuji orchard encountered significant wind and rain in the evening on the date of the field trial, which could have had a severe impact on the observed fruit set outcomes.

\subsection{Cycle Time}
The average cycle time for pollinating individual flower clusters for the proposed robotic pollination system was 6.5 seconds. Table \ref{tab:cycletime_breakdown} shows the breakdown of the time required for tasks involved to complete a successful spray of pollens to the target flowers. As shown in Table \ref{tab:cycletime_breakdown}, the manipulator execution and the spray time were the most time-consuming tasks, with manipulator execution taking about 50\% and spray time taking 30\% of total cycle time. Hence, optimizing the manipulator execution and the spray time could potentially reduce the overall system cycle time. This optimization could be achieved using linear telescoping manipulators instead of the complex UR5e robotic manipulator with 6-DOF. Linear telescoping manipulators are simpler to control and can be programmed to move more quickly and efficiently. Additionally, a more comprehensive investigation of the sprayer system and the spraying mechanism could help improve the efficient pollen delivery. This could involve developing new sprayer nozzle designs or using new control algorithms to optimize the spraying process.
Furthermore, a more powerful computing system coupled with more efficient detection, segmentation, and pose estimation software could help reduce the time required for machine vision tasks.  

\begin{table}[]
    \centering
    \begin{tabular}{lc}
        \hline \hline
         Tasks & Time (seconds)  \\
         \hline
         Segmentation & 0.4 \\
         Pose estimation &0.8 \\
         Motion plan computation &0.1 \\
         Manipulator execution &3.2 \\
         Pollen spray &2.0 \\
         \hline
         Total & 6.5 \\
         \hline \hline
    \end{tabular}
    \caption{Breakdown of average cycle time required to pollinate individual flower clusters. The majority of the cycle time was used to complete manipulator execution and spraying of pollens.}
    \label{tab:cycletime_breakdown}
\end{table}

\subsection{Fruit Quality Assessment}
\begin{figure}[!ht]
     \centering
     \includegraphics [width=0.99\textwidth]{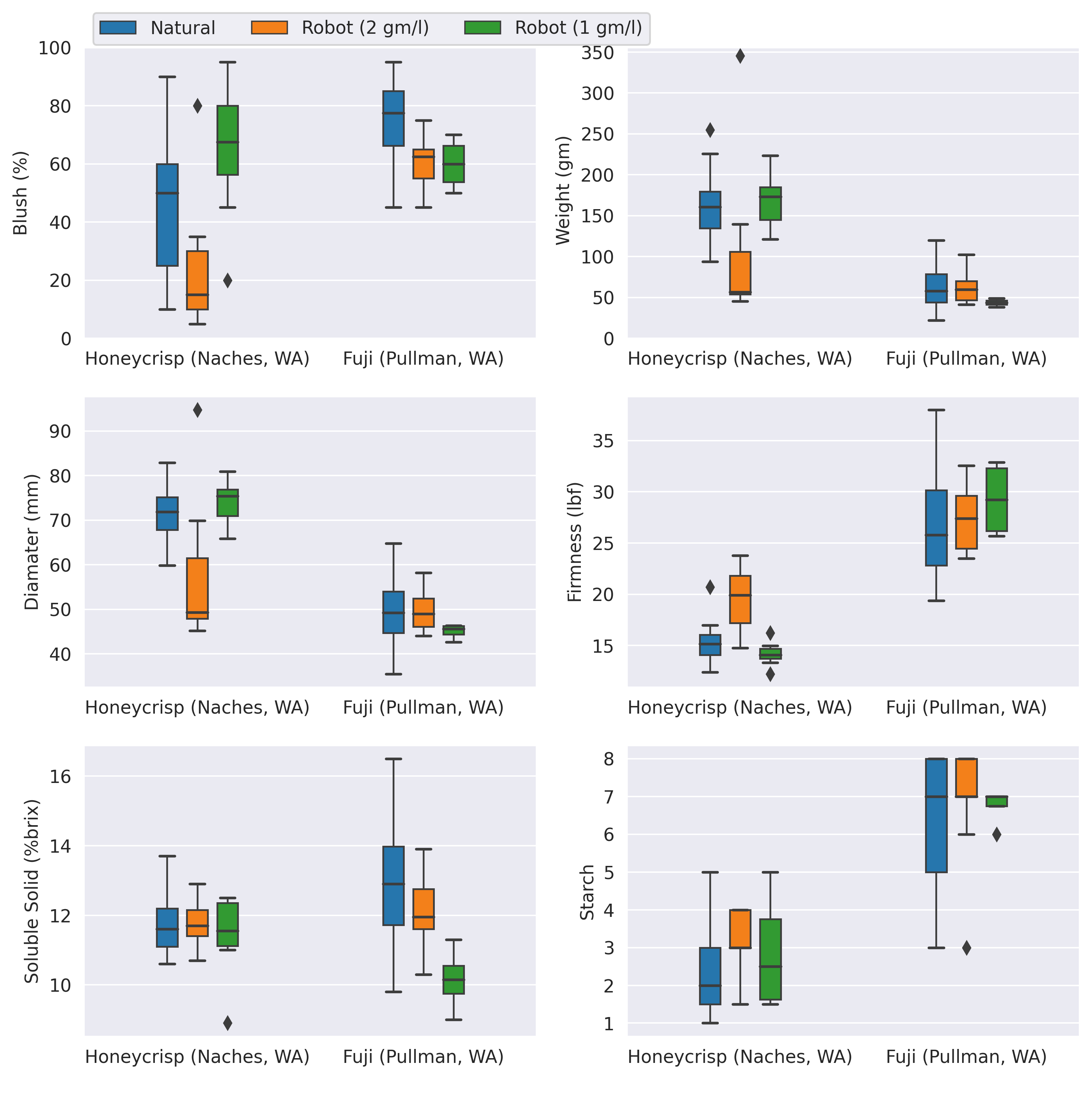}
     \caption{Comparison of external fruit quality parameters of harvested Honeycrisp and Fuji apples pollinated via robotic and natural pollination approaches. Robotic pollination consistently demonstrated comparable and, at times exhibited superior fruit quality compared to natural pollination.}
     \label{fig:quality_comparison}
 \end{figure}

The apple orchards were visited during the harvest season, and fruits from the experimental clusters were harvested for fruit quality evaluation. Figure \ref{fig:pollination_harvested} shows robotically pollinated harvest-ready Honeycrisp and harvested Fuji apples during the fruit quality assessment. As discussed in Section \ref{FruitsetandQuality}, quality of fruit produced with robotic and natural pollination methods was evaluated using six internal and external fruit quality metrics with the results shown in Table \ref{tab:fruitquality_evaluation}, and Figure \ref{fig:quality_comparison}.
% While Fuji apples showed a lower fruit set compared to the Honeycrisp, the quality of the robotically pollinated Fuji apples was comparable to the natural pollination.

\label{sec:result_sizecolor}
\begin{table}[ht]
    \centering
    \caption{Fruit quality comparison of robotic vs natural pollination}
    \label{tab:fruitquality_evaluation}
   \scalebox{0.81}{
   \begin{tabular}{ccccccccc}
        \hline  \hline
        Experiment Site &Pollination & Color  &Weight  &Diameter &Firmness &Soluble Solid  &Starch \\
                        &Approach    & (\%blush) & (gm) &(mm)    &(lbf)    &(\% Brix)      & Scale\\
        \hline
         Honeycrisp     & Natural       &46$\pm$21 &161.0$\pm$39.2 &72.0$\pm$6.2 &15.2$\pm$1.6 &11.7$\pm$0.9 &3.3$\pm$1.2 \\
         (Naches, WA)   & Robot (2 gm/l) &23$\pm$22$^*$ &98.1$\pm$88.6$^*$ &56.9$\pm$15.1$^*$ &\textbf{19.6$\pm$3.3}$^*$ &\textbf{11.8$\pm$0.6}  &3.2$\pm$1.0  \\
                        & Robot (1 gm/l) &\textbf{66$\pm$22}$^*$ &\textbf{169.4$\pm$31.3} &\textbf{74.3$\pm$4.6} &14.2$\pm$1.1 &11.5$\pm$1.1 &\textbf{3.6$\pm$1.0} \\
                        % &\cline{1-8}
                        % & Significant differences &$N>R_2,N<R_1,R_2<R_1$\\
        \hline
        Fuji           & Natural        &\textbf{74$\pm$15} &\textbf{66.5$\pm$25.6} &48.4$\pm$7.0 &26.3$\pm$4.8 &\textbf{12.6$\pm$1.8}  &{6.4$\pm$1.5} \\
        (Pullman, WA)  & Robot (2 gm/l)  &61$\pm$10$^*$ &61.7$\pm$19.5 &\textbf{49.5$\pm$4.6} &27.4$\pm$3.3 &12.1$\pm$1.2 &\textbf{6.9$\pm$1.6} \\
                       & Robot (1 gm/l)  &60$\pm$9$^*$ &43.8$\pm$4.6 &45.1$\pm$1.7 &\textbf{29.2$\pm$3.8} &10.2$\pm$1.0$^*$ &{6.7$\pm$0.5} \\
                       % & Robot (0.5 gm/l) &\textbf{77$\pm$16} &\textbf{76.1$\pm$49.1} &49.4$\pm$4.1 &26.5$\pm$4.6 &12.4$\pm$1.8 &{6.7$\pm$1.3}  \\
                       % &\cline{1-8}
                       % & Significant differences \\
       
        \hline \hline
    \end{tabular}}
    
\end{table}

\subsubsection{Color, Weight and Diameter}
For the Honeycrisp variety, the robotic pollination approach at 1 gm/l pollen concentration produced fruits with significantly higher blush [Mean: 66\%, 95\% Confidence Interval (CI): 50\% to 81\%, \textit{p $<$ 0.05}], compared to those harvested from natural pollination [Mean: 46\%, 95\% CI: 38\% to 54\%] (see Table \ref{tab:fruitquality_evaluation}). On the other hand, robotic pollination at 2 gm/l pollen concentration produced fruits with significantly lower fruit blush [Mean: 23\%, 95\% CI: 9\% to 38\%, \textit{p $<$ 0.05}] when compared to naturally pollinated fruits. Additionally, the robotic pollination with 1 gm/l pollen concentration produced larger fruits with higher weight [Mean: 169.4, 95\% CI: 147.0 to 191.9, \textit{p $>$ 0.05}] and diameter  [Mean: 74.3, 95\% CI: 71.0 to 77.6, \textit{p $>$ 0.05}] compared to naturally pollinated fruits [Weight: Mean: 161, 95\% CI: 144.8 to 177.2; Diameter: Mean: 72.0, 95\% CI: 69.5 to 74.6], but these differences were not statistically significant. Conversely, 2 gm/l pollen concentration produced significantly smaller fruits with lower weight [Mean: 98.1, 95\% CI: 38.6 to 157.6, \textit{p $<$ 0.05}] and diameter [Mean: 56.9, 95\% CI: 46.74 to 67.04, \textit{p $<$ 0.05}] which also exhibited shape disorder. We believe the netting structure used to prevent natural pollination may have contributed to the reduced fruit size and shape disorders observed in the 2 gm/l pollen concentration treatment. During the assessment of the fruit set, it was observed that the netting structure for some of the flowers treated with 2 gm/l pollen concentration was severely distorted. This deformation may have placed stress on the leaves, netted branches, and fruitlets, potentially inhibiting proper fruit growth and leading to physiological abnormalities.

For the Fuji apple variety, the robotic pollination at 2 gm/l pollen concentration [Mean: 61\%, 95\% CI: 54\% to 68\%, \textit{p $<$ 0.05}] and 1 gm/l pollen concentration [Mean: 60\%, 95\% CI: 46\% to 75\%, \textit{p $<$ 0.05}] produced significantly lower fruit blush compared to the natural pollination [Mean: 74\%, 95\% CI: 70\% to 81\%]. Furthermore, robotic approach with 2 gm/l pollen concentration yielded larger fruits with greater weight [Mean: 61.7, 95\% CI: 48.3 to 75.1] and diameter [Mean: 49.5, 95\% CI: 46.3 to 52.7] compared to 1 gm/l pollen concentration [Weight: Mean: 43.8, 95\% CI: 36.4 to 51.1; Diameter: Mean: 45.1, 95\% CI: 42.3 to 47.8]. The statistical analysis did not show significant differences in fruit weight and diameter among the fruits produced by both robotic pollination approaches when compared against natural pollination (\textit{p $>$ 0.05}). The inter-varietal comparison showed that the produced Honeycrisp apples were bigger with higher weight and diameter, and exhibited a lower color blush than Fuji apples for both robotic and natural pollination approaches.

\subsubsection{Firmness, Soluble Solid, and Starch Scale}
As the fruits mature, their firmness decreases, and sugar content increases since starch gets hydrolyzed into sugar. For the Honeycrisp variety, fruits produced from the 1 gm/l pollen concentration were significantly less firm [Mean: 14.2, 95\% CI: 13.6 to 14.8, \textit{p$<$0.05}], signifying more maturity than those resulting from natural pollination [Mean: 15.2, 95\% CI: 14.7 to 15.7] (see Table \ref{tab:fruitquality_evaluation}). Conversely, fruits produced from 2 gm/l pollen concentration fruits were significantly firmer [Mean: 19.6, 95\% CI: 18.1 to 21.0, \textit{p$<$0.05}] than those from natural pollination. On the other hand, the soluble solid contents from both 1 gm/l [Mean: 11.5, 95\% CI: 10.7 to 12.3, \textit{p $>$ 0.05}] and 2 gm/l [Mean: 11.8, 95\% CI: 11.4 to 12.2, \textit{p $>$ 0.05}] pollen concentration showed slightly higher but statistically insignificant results compared to natural pollination [Mean: 11.7, 95\% CI: 11.4 to 12.1]. Furthermore, the starch iodine test showed fruits pollinated with 1 gm/l pollen concentration showed higher ripeness [Mean: 3.6, 95\% CI: 3.0 to 4.2, \textit{p $>$ 0.05}]  than those resulting from natural pollination [Mean: 3.3, 95\% CI: 2.8 to 3.7] and robotic pollination with 2 gm/l pollen concentration [Mean: 3.2, 95\% CI: 2.5 to 3.7, \textit{p $>$ 0.05}]. However, these differences were also not statistically significant.

For the Fuji apple variety, the naturally pollinated fruits were the least firm [Firmness: Mean: 26.3, 95\% CI: 25.6 to 28.2] and had the highest soluble solid content [Soluble Solid: Mean: 12.6, 95\% CI: 12.4 to 13.7] compared to both robotic pollination approaches with no significant differences compared to fruits produced from 2 gm/l pollen concentration[Firmness: Mean: 27.4, 95\% CI: 25.7 to 29.0, \textit{p $>$ 0.05}; Soluble Solid: Mean: 12.1, 95\% CI: 11.3 to 12.9, \textit{p $>$ 0.05}] (see Table \ref{tab:fruitquality_evaluation}). On the other hand, while the fruit firmness [Mean: 29.2, 95\% CI: 26.6 to 32.4, \textit{p $>$ 0.05}] from the 1 gm/l pollen concentration did not show statistical significance compared to naturally pollinated fruits, the soluble solid content was lowest [Mean: 10.2, 95\% CI: 8.6 to 11.7, \textit{p $<$ 0.05}] and exhibited statistical significance. Furthermore, the starch iodine test showed comparative results among the pollination approaches with fruits pollinated with 2 gm/l pollen concentration showing slightly higher ripeness [Mean: 6.9, 95\% CI: 5.9 to 7.9, \textit{p $>$ 0.05}]   followed by robotic pollination with 1 gm/l pollen concentration [Mean: 6.7, 95\% CI: 6.2 to 7.2, \textit{p $>$ 0.05}], and natural pollination [Mean: 6.4, 95\% CI: 5.9 to 6.9]. 

Evaluation of internal fruit qualities such as firmness, soluble solids, and starch scale suggest that robotic pollination can influence the ripening process of both Honeycrisp and Fuji apples. The inter-varietal comparison showed that Honeycrisp apples were less firm and had more consistent soluble solid content compared to Fuji apples. The robotic pollination with 1 gm/l pollen concentration increased the ripeness of Honeycrisp apples, as evidenced by their lower firmness and higher starch iodine test scores. For Fuji apples, robotic pollination produced results that were comparable with natural pollination, with minimal numerical differences and no statistically significant differences in firmness, soluble solids content, or starch scale.

In summary, the results suggest the complex interactions between pollen concentration, pollination methods, apple variety, and environmental factors may play a vital role in achieving the desired fruit set and quality. Natural pollination has been the primary source of fruit production for centuries and remains the most reliable pollination approach. The results show that robotic pollination approaches, especially at high pollen concentration, could be a viable alternative to natural pollination, providing fruit quality comparable to natural pollination.  However, the results may be variety dependant. Further research is needed to confirm the findings in this study and to determine the optimal robotic pollination parameters for different apple varieties.

% As shown in  Table \ref{tab:fruitquality_evaluation} and Figure \ref{fig:quality_comparison}, natural pollination resulted in the highest average seed count for both Honeycrisp (seed count: 3.9) and Fuji (seed count: 5.8) apples. However, the seed count for robotic pollination approaches for Honeycrisp was substantially lower(1.0 for 2 gm/l and 1.9 for 1 gm/l), without any seeds in the fruit seed pockets of some of the apples. Conversely, the robotic pollination methods yielded promising seed counts across all three treatment approaches for Fuji apples. Among the robotic pollination methods, the 2 gm/l pollen concentration yielded the same average seed count as natural pollination (5.8), followed by the 1 gm/l (4.5) and 0.5 gm/l (4.7) pollen concentrations.

%%%%%%%%%%%%%%% Naches %%%%%%%%%%%%%%%%%%%%%%%%%%
% Seed Count
% Natural [Mean:3.9, 95\% CI:3.0 to 4.8]
% 2 gm/l [Mean:1.0, 95\% CI:-0.8 to 2.8]
% 1 gm/l [Mean:1.9, 95\% CI: 0.1 to 3.7]

%%%%%%%%%%%%%%%%%%%%%%%%%%%%Pullman%%%%%%%%%%%%%%%%%%%%%%%%%%%%%%%%
% Seed count
% Natural [Mean: 6.4, 95\% CI: 5.5 to 7.3]
% 2 gm/l [Mean: 5.8, 95\% CI: 4.0 to 7.6]
% 1 gm/l [Mean: 4.5, 95\% CI: 1.7 to 7.3]
% 0.5 gm/l [Mean: 4.7, 95\% CI: 3.6 to 5.8]

\subsection{Physiological Disorders}
\begin{figure}[ht]
    \centering
    \includegraphics[width=0.35\textwidth]{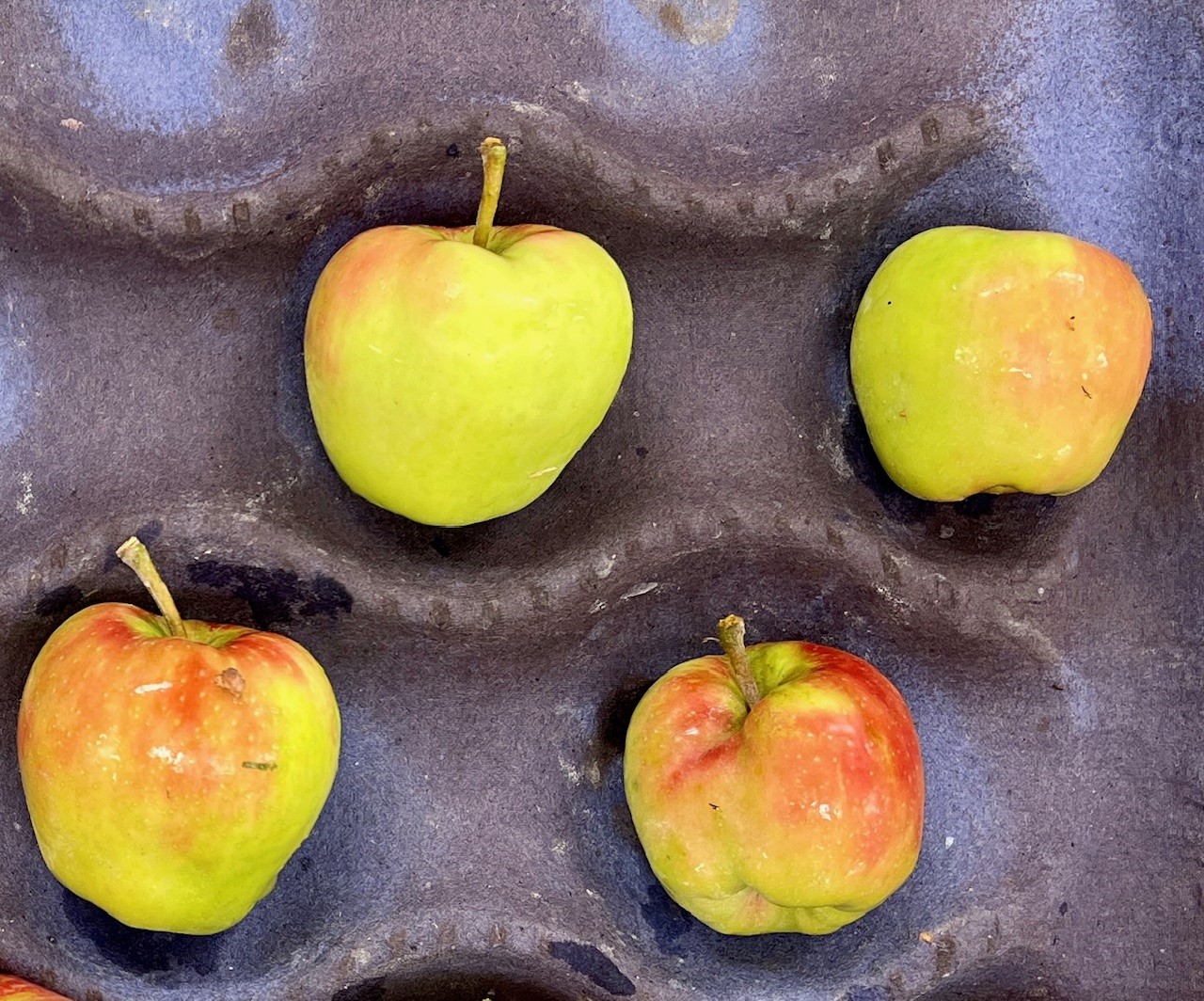}
    \includegraphics[width=0.376\textwidth]{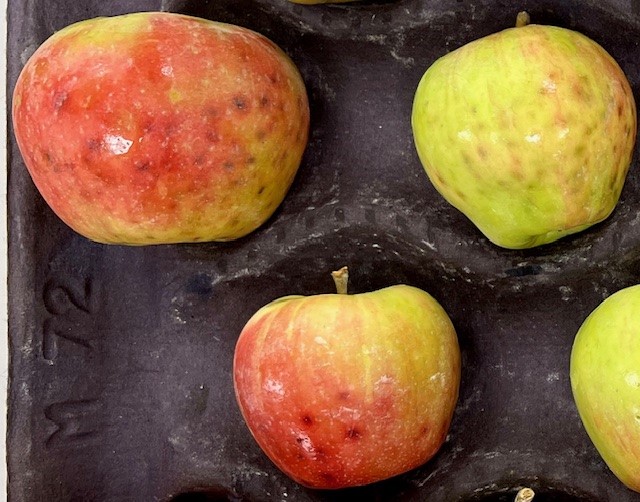}
    \includegraphics[width=0.365\textwidth]{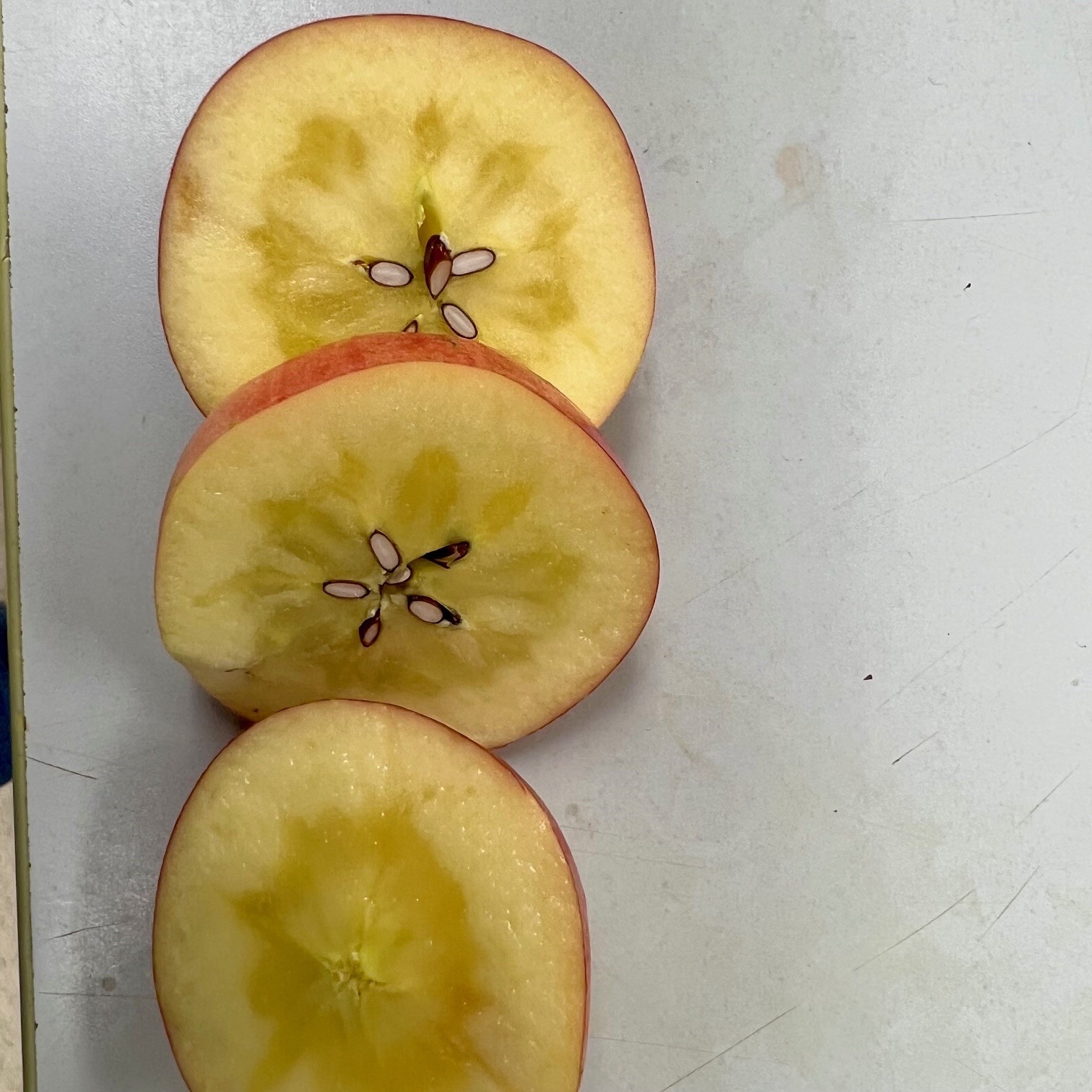}
    \includegraphics[width=0.365\textwidth]{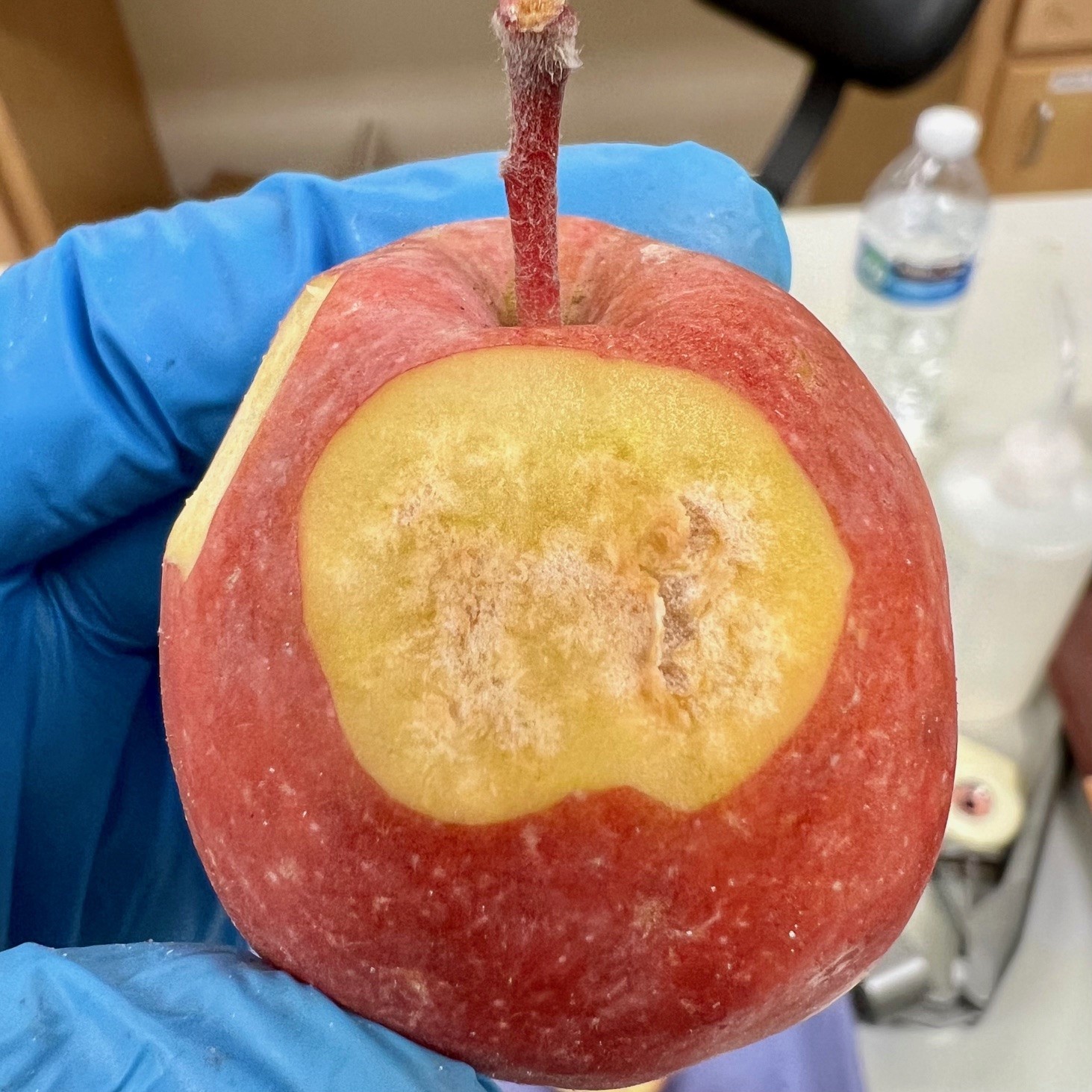}
    
    \caption{Physiological disorder in Honeycrisp and Fuji apples. Shape and size disorder in Honeycrisp observed in robotic pollination experiment with 2 gm/l pollen concentration [Upper Left], Bitterpit disorder in Honeycrisp [Upper Right], Watercore disorder in Fuji [Lower Left], physiological disorder in Fuji [Lower Right]}
    \label{fig:physicaldisorder}
\end{figure}

Different physiological disorders were observed in both apple varieties during fruit harvest and quality evaluation. Some disorders were specific to a particular pollination approach, while others were unique to the cultivar and were observed in fruits produced from all pollination approaches. For instance,  Honeycrisp apples from Naches, WA, pollinated with a 2 gm/l pollen concentration, were smaller and deformed in shape, likely due to the netting structure (Figure \ref{fig:physicaldisorder}[Upper Left]) (also discussed in Section \ref{sec:result_sizecolor}). Additionally, there were instances of fruits displaying symptoms of bitter pit disorder in Honeycrisp apples, irrespective of pollination approach (Figure \ref{fig:physicaldisorder}[Upper Right]). On the other hand, some of the Fuji apples across all pollination approaches showed signs of watercore (Figure \ref{fig:physicaldisorder}[Lower Left]) and another physiological disorder (Figure \ref{fig:physicaldisorder}[Lower Right]). It is important to note that physiological disorders in fruit can be caused by various factors, including genetics, environmental conditions, and nutritional imbalance \citep{watkins2019apple}, which were out of scope of this work.

\subsection{Limitations}
This study introduces a new approach to robotic pollination of apples, but it has some limitations that need further research. One area for improvement is the relatively high cycle time of 6.5 seconds per flower cluster. This could be addressed by using simpler linear robotic manipulators and improving motion planning algorithms to enhance system efficiency. It is also important to investigate the optimal spray duration, as the current 2-second spray time may not be ideal for maximizing pollination effectiveness while minimizing cycle time. The small sample sizes, especially for robotic pollination treatments, could impact the statistical power and generalizability of the findings, so larger-scale studies across different apple varieties are needed to validate the system's viability. Additionally, while the pollen suspension was agitated before and during application through recirculation, the effects of continuous agitation on pollen viability and suspension consistency require further examination. Environmental factors, such as weather conditions and orchard architecture, were not fully controlled in this study, potentially influencing pollination outcomes. Future research should also address the long-term impacts of robotic pollination on tree health, fruit quality, and yield over multiple growing seasons, as well as conduct comprehensive cost-benefit analyses to assess the economic feasibility of widespread adoption in commercial apple production.

\section{Conclusion}
In this study, a robotic pollination system was designed, developed, integrated, and tested in a Honeycrisp and Fuji apple orchards. The robotic system comprised a stereo vision sensing, deep learning-based image processing, robotic manipulation, and an electrostatic sprayer system for precise spray of pollens to the target flower clusters. The effectiveness of the robotic pollination approach was evaluated by comparing the fruit set and quality of the harvested fruits with the same from natural pollination. Based on the results of this study, the following conclusions were drawn:

\begin{itemize}
    \item Deep learning-based machine vision system showed promising results in identifying and segmenting flower clusters in complex orchard images. The machine vision system achieved a mean average precision of 0.89 in delineating target cluster boundaries.
    
    \item  Robotic pollination has shown to be effective in natural apple orchard environments. In Honeycrisp apples, using a pollen concentration of 2 gm/l resulted in fruit set in 87.5\% of clusters, slightly lower than the 94.9\% fruit set achieved through natural pollination. However, for Fuji apples, the same pollen concentration only resulted in fruit set for 20.6\% of sprayed flower clusters. This suggests that the effectiveness of robotic pollination may be influenced by factors such as cultivar characteristics, tree canopy structure, orchard age, weather conditions during pollination, and other environmental variables. Despite the difference in fruit set, the quality of fruits produced through robotic pollination was comparable to those resulting from natural pollination.
    
\end{itemize}

The proposed robotic pollination system offers potential for improvement through advancements in machine vision, mechatronic systems, fruit set capabilities, and fine-tuning of pollen spray parameters like duration and distance. A balance between desired fruit set and yield needs to be studied, especially considering potential fruitlet thinning requirements.  This study, while having limitations, represents one of the first trials of robotic pollination in an outdoor apple orchard environment. It provides a promising foundation for developing robotic technologies as an alternative pollination strategy, contributing to the long-term sustainability of fruit crop production.

\subsubsection*{Acknowledgments}
This research is partially funded by award No. The authors express sincere gratitude to Dave Allan of Allan Bros. Inc. for generously providing access to his commercial apple orchard in Naches, WA for field evaluations. We also extend our appreciation to Deb Pehrson for facilitating access to the Washington State University research orchard in Pullman, WA during data collection and field trials.
\bibliographystyle{apalike}
% \bibliography{jfrExampleRefs}

\end{document}